\documentclass[11pt]{article}

\usepackage[utf8]{inputenc}
\usepackage[T1]{fontenc}
\usepackage{amsmath,amssymb,amsfonts}
\usepackage{graphicx}
\usepackage{booktabs}
\usepackage{multirow}
\usepackage{makecell}
\usepackage{xcolor}
\usepackage{hyperref}
\usepackage{cleveref}
\usepackage{enumitem}
\usepackage{natbib}
\usepackage[margin=1in]{geometry}
\usepackage{tikz}
\usetikzlibrary{trees,positioning,arrows.meta,shapes.geometric}
\usepackage[edges]{forest}
\usepackage{tcolorbox}
\tcbuselibrary{skins,breakable}
\usepackage{colortbl}
\usepackage{tabularx}
\usepackage{adjustbox}
\usepackage{placeins}

\hypersetup{
  colorlinks=true,
  linkcolor=blue!55!black,
  citecolor=green!45!black,
  urlcolor=cyan!45!black,
  pdfborder={0 0 0},
  pdftitle={From Reasoning to Agentic: Credit Assignment in Reinforcement Learning for Large Language Models},
  pdfauthor={Chenchen Zhang}
}
\newcommand{\method}[1]{\textsc{#1}}

\newtcolorbox{insightbox}[1][]{
  colback=blue!5!white,
  colframe=blue!60!black,
  fonttitle=\bfseries,
  title={#1},
  rounded corners,
  boxrule=0.5pt,
  left=4pt, right=4pt, top=4pt, bottom=4pt,
  before skip=8pt, after skip=8pt,
}
\newtcolorbox{challengebox}[1][]{
  colback=red!4!white,
  colframe=red!60!black,
  fonttitle=\bfseries,
  title={#1},
  rounded corners,
  boxrule=0.5pt,
  left=4pt, right=4pt, top=4pt, bottom=4pt,
  before skip=8pt, after skip=8pt,
}
\newtcolorbox{takebox}[1][]{
  colback=green!4!white,
  colframe=green!50!black,
  fonttitle=\bfseries,
  title={#1},
  rounded corners,
  boxrule=0.5pt,
  left=4pt, right=4pt, top=4pt, bottom=4pt,
  before skip=8pt, after skip=8pt,
}

\title{From Reasoning to Agentic: Credit Assignment\\in Reinforcement Learning for Large Language Models}

\author{
  Chenchen Zhang \\
  Independent Researcher \\
  \texttt{zcc1959339538@gmail.com}
}

\date{}

\begin{document}
\maketitle

\begin{abstract}
Reinforcement learning (RL) for large language models (LLMs) increasingly relies on sparse outcome rewards, yet those rewards say little about which token, reasoning step, tool call, memory operation, or agent caused an outcome. This \emph{credit assignment} (CA) problem spans reasoning RL and becomes sharper in agentic RL, where environment interaction introduces transition non-closure, partial observability, limited replay, heterogeneous actions, weak intermediate verifiability, and agent coupling.

We synthesize a unified corpus of 69 papers published from January 2024 through July 31, 2026: 56 core CA methods and 13 adjacent or boundary enablers, selected from 92 deduplicated screening records. We retain the original granularity $\times$ methodology taxonomy and add a six-diagnostic framework that maps assumption breaks to identification barriers, estimators, and evaluation controls. A source-located full-text audit covers a fixed 42-core-paper subset. Two algorithm researchers independently and blindly cross-coded its 252 diagnostic cells, agreeing on 223 (88.5\%); per-diagnostic Cohen's $\kappa$ ranges from .543 to .909, and principal-family agreement is 42/42 ($\kappa=1.000$).

Beyond taxonomy, we establish when restored-state comparisons identify a protocol-specific causal contrast, show that text-only histories can leave even the sign of credit unidentified, and introduce a reusable CA-ID Card linking each claim to its estimand, evidence provenance, and falsification test. An atomic reporting audit describes comparator, budget-parity, ablation, overhead, uncertainty, and replay coverage without constructing a cross-paper leaderboard. The companion repository hosts a living catalog and decision aids at \url{https://github.com/xxzcc/Awesome-Credit-Assignment-in-LLM-RL}; a dated, version-specific release of the frozen audit bundle is planned there separately from the minimal arXiv source.
\end{abstract}

\section{Introduction}
\label{sec:intro}

The past two years have witnessed two waves of reinforcement learning for large language models. The first wave---\textbf{reasoning RL}---demonstrated that RL can dramatically improve LLMs' ability to solve mathematical problems, write code, and perform logical reasoning~\citep{deepseekr1,shao2024grpo}. Models like DeepSeek-R1 and OpenAI's o1 showed that training with outcome-level rewards (``is the final answer correct?'') can elicit sophisticated chain-of-thought reasoning. The second wave---\textbf{agentic RL}---extends this paradigm to multi-turn interactive tasks: LLM agents that browse the web~\citep{zhou2024webarena}, use tools~\citep{schick2024toolformer}, write and debug code, and collaborate with other agents. This shift from reasoning to agency represents a qualitative leap in the complexity of the RL problem.

At the heart of both waves lies a shared bottleneck: \textbf{credit assignment}. When the only feedback is a sparse terminal reward (``problem solved'' or ``task completed''), how do we determine which specific actions---which tokens, which reasoning steps, which tool calls---were responsible?

\paragraph{Why credit assignment is the core bottleneck.} The severity of the credit assignment problem scales with trajectory complexity:

\begin{itemize}[nosep]
    \item In \textbf{reasoning RL}, a typical trajectory is a single LLM generation ranging from $\sim$500 tokens (GSM8K-level problems) to 10,000--30,000+ tokens for hard competition mathematics (e.g., DeepSeek-R1 averages $\sim$23K tokens on AIME 2025~\citep{deepseekr1}). Credit must be distributed across tokens and reasoning segments. Episode-level methods like GRPO~\citep{shao2024grpo} and REINFORCE assign the same advantage to every token---a crude approximation that nonetheless works for shorter trajectories.
    \item In \textbf{agentic RL}, trajectories span 10--100+ turns, each involving an LLM call plus environment interaction. The total token count routinely reaches 100K--500K+ (e.g., in one reported SWE-bench setup, agents averaged $\sim$64 turns consuming $\sim$131K tokens~\citep{ragen}). Episode-level credit becomes increasingly uninformative: a single wrong tool call at turn 3 receives the same penalty as dozens of correct subsequent actions.
\end{itemize}

The community has responded with a burst of innovation. Our July 31, 2026 snapshot contains 69 included papers---56 core CA methods and 13 adjacent or boundary enablers---ranging from Monte Carlo token-level value estimation~\citep{vineppo} to Shapley value-based reward decomposition~\citep{scar,sharp}, and from process reward models~\citep{agentprm,pure} to hindsight counterfactual analysis~\citep{hcapo,c3,ccpo}. The corpus contains 21 reasoning, 31 agentic or mixed, and 4 multi-agent core methods, making the reasoning-to-agentic shift visible in both mechanism design and evaluation requirements.

\paragraph{Scope and inclusion criteria.} We include methods whose \emph{primary contribution} is a novel credit assignment mechanism for LLM RL. We distinguish between \textbf{core CA methods}---which directly change how an outcome-derived learning signal is assigned across actions---and \textbf{adjacent or boundary enablers}, where reward shaping, exploration, skill distillation, segmentation, infrastructure, or evaluator design supplies a closely related signal without directly estimating outcome responsibility. Both categories remain in the 69-paper qualitative corpus, but only the fixed 42-paper full-text subset supports the diagnostic reliability and reporting-prevalence statistics below.

\paragraph{Scope and narrative.} Unlike existing work that treats credit assignment as a sub-topic~\citep{zhang2025landscape} or focuses on classical RL~\citep{pignatelli2023survey}, this paper makes credit assignment the central lens through which we examine LLM RL. Our narrative arc is:

\begin{center}
\emph{Classical RL} $\rightarrow$ \emph{Reasoning RL} $\rightarrow$ \emph{Agentic RL} $\rightarrow$ \emph{Future: Multi-Agent Systems}
\end{center}

At each stage, the credit assignment problem grows harder, and new methods emerge to meet the challenge.

\paragraph{Contributions.} This paper makes three distinct types of contribution:

\medskip\noindent\textit{I. Survey with taxonomy.}
\begin{enumerate}[nosep]
    \item \textbf{Dedicated analysis}: We provide a dedicated survey focused on credit assignment in LLM RL, covering both reasoning and agentic settings (\Cref{sec:reasoning_ca,sec:agentic_ca}).
    \item \textbf{Two complementary views}: We organize 69 papers by \emph{granularity} $\times$ \emph{methodology} and map six diagnostic breaks to identification barriers, estimators, and evaluation controls (\Cref{sec:taxonomy,sec:diagnostics}).
    \item \textbf{Reasoning $\to$ Agentic analysis}: We explicitly characterize \emph{why} agentic RL makes credit assignment qualitatively harder and what new techniques this demands (\Cref{sec:why_agentic_harder}).
    \item \textbf{Identification boundary}: We state a restored-state identification result, give a constructive text-only non-identifiability witness, and separate restored-interventional, observational-causal, model-relative, and predictive/proxy claims (\Cref{sec:identification}).
\end{enumerate}

\medskip\noindent\textit{II. Reusable structured artifact.}
\begin{enumerate}[nosep,resume]
    \item \textbf{Auditable corpus and coding}: We construct a 69-paper inventory, a 92-record screening ledger, source-located diagnostic labels, two frozen blind coding ledgers, and atomic reporting fields for a 42-core-paper subset (\Cref{sec:methodology,app:inventory,app:checklist}).
\end{enumerate}

\medskip\noindent\textit{III. Standardization proposals.}
\begin{enumerate}[nosep,resume]
    \item \textbf{CA-ID Card and reporting audit}: We propose a machine-readable claim contract and audit comparator, budget parity, ablation, uncertainty, overhead, and replay provenance without treating heterogeneous paper results as a leaderboard (\Cref{sec:identification,app:checklist}).
    \item \textbf{Benchmark protocol}: We outline a minimal specification for a credit assignment evaluation suite, including task families, required metadata, and controlled bifurcation tasks (\Cref{sec:open}).
    \item \textbf{Research roadmap}: We identify open problems at the frontier---multi-agent credit, ultra-long horizons, the exploration-credit interplay---and identify agentic RL as a likely driver of future innovation (\Cref{sec:open}).
\end{enumerate}

\paragraph{Relation to existing work.} \citet{pignatelli2023survey} provide an excellent review of temporal credit assignment in classical deep RL (56 pages, 2023), but predate the LLM era entirely. \citet{zhang2025landscape} offer a comprehensive 100-page overview of agentic RL for LLMs (500+ papers), but treat credit assignment as one sub-topic among many without depth. Several works on reasoning RL~\citep{rlforlrm} cover RL algorithms broadly but do not focus on credit assignment. To the best of our knowledge, no existing work systematically examines credit assignment across both reasoning and agentic LLM RL.

\paragraph{Paper organization.} \Cref{sec:background} introduces the background, problem formulation, and taxonomy. \Cref{sec:reasoning_ca} reviews reasoning methods, \Cref{sec:why_agentic_harder,sec:agentic_ca} analyze agentic settings, and \Cref{sec:multi_agent} covers multi-agent credit. \Cref{sec:frontier_update} integrates the July update, while \Cref{sec:comparison} provides a structured legacy comparison and reporting audit. \Cref{sec:pipeline,sec:open} discuss pipeline interactions and open problems before \Cref{sec:conclusion} concludes.

\paragraph{How to use this survey.} This paper is designed to serve different readers in different ways:
\begin{itemize}[nosep]
    \item \textbf{Practitioners} choosing a CA method for a specific task: start with the decision tree (\Cref{fig:decision_tree}) and recommendation table (\Cref{tab:recommendation}), then read the relevant method section for details.
    \item \textbf{Researchers} seeking open problems: read \Cref{sec:why_agentic_harder} for the core challenges, then \Cref{sec:open} for the research roadmap. The benchmark protocol (\Cref{sec:open}) and reporting checklist (\Cref{app:checklist}) may inform experimental design.
    \item \textbf{Reviewers and meta-researchers}: the structured inventory (\Cref{app:inventory}) provides machine-readable metadata for all 69 papers; the source-located 42-paper audit (\Cref{app:checklist}) documents current reporting gaps and coding uncertainty.
    \item \textbf{Newcomers} to LLM RL credit assignment: read \Cref{sec:background} for foundations, then follow the narrative arc through \Cref{sec:reasoning_ca,sec:agentic_ca}.
\end{itemize}

\subsection{Literature Coverage}
\label{sec:methodology}

This survey covers January 2024 through July 31, 2026. Searches combined CA terms (``credit assignment,'' ``process reward,'' ``reward decomposition,'' ``turn-level reward,'' ``advantage redistribution'') with LLM/RL and agent terms across arXiv, Semantic Scholar, Google Scholar, venue proceedings, and citation chains. After deduplication, the screening ledger contains 92 records. Applying explicit core, adjacent/boundary, related, evaluation-only, and exclusion rules yields a unified corpus of 69 included papers: 56 core methods and 13 adjacent/boundary enablers. The remaining 23 records stay in the ledger with reasons rather than disappearing from the denominator.

Corpus-level taxonomy and qualitative synthesis use all 69 included papers. A fixed 42-core-paper subset---15 reasoning, 23 agentic or mixed, and 4 multi-agent---supports full-text diagnostic, inter-coder, and reporting-prevalence analyses. It is not a random sample of the 56 core methods, so no 42-paper statistic is presented as corpus-wide prevalence. Under a rubric frozen before recoding, two algorithm researchers independently and blindly coded the six diagnostics and one exclusive principal family. They agreed on 223/252 binary cells (88.5\%), with per-diagnostic Cohen's $\kappa$ from .543 to .909, and on all 42 family labels ($\kappa=1.000$). The 29 diagnostic disagreements across 18 papers remain preserved rather than post-hoc reconciled into the reliability numbers.

This is a structured, auditable framework survey, not a systematic review: query selection and boundary coding remain judgment-dependent, rapid preprint turnover may create recall gaps, and the fixed full-text subset limits prevalence claims. The companion repository provides the living catalog and decision aids. A dated repository release of the complete frozen inventory, screening ledger, query protocol, coding rubric and ledgers, reporting audit, and blank CA-ID Card is planned separately from the minimal arXiv source; \Cref{sec:threats} states the corresponding validity limits.

\section{Background and Problem Formulation}
\label{sec:background}

\subsection{From Reasoning RL to Agentic RL: A Brief History}
\label{sec:history}

The application of RL to LLMs has progressed through several distinct phases, each introducing new credit assignment challenges.

\paragraph{Phase 1: RLHF (2022--2023).} InstructGPT~\citep{ouyang2022rlhf} and its successors established the paradigm of training a reward model from human preferences and fine-tuning the LLM via PPO. In this setting, trajectories are single-turn responses of moderate length ($\sim$500 tokens), and the reward model provides a \emph{dense} scalar signal for the entire response. Credit assignment is implicit: PPO's learned value function provides token-level baselines, though the quality of these baselines in the high-dimensional LLM action space remains debated.

\paragraph{Phase 2: Reasoning RL (2023--2025).} A breakthrough emerged when researchers discovered that training LLMs with RL on \emph{verifiable} outcome rewards---without any reward model---could elicit sophisticated reasoning behavior. DeepSeek-R1~\citep{deepseekr1} demonstrated that GRPO with binary correctness rewards on mathematical problems produces models capable of extended chain-of-thought reasoning. OpenAI's o1 and o3 models showed similar capabilities. In this phase, trajectories are single generations ranging from $\sim$500 tokens (easy math) to 30,000+ tokens (hard competition problems; DeepSeek-R1 averages $\sim$23K tokens on AIME~\citep{deepseekr1}), and the reward is purely terminal (correct or incorrect). Credit assignment becomes explicit: how should the single outcome reward be distributed across thousands of reasoning tokens? This question spawned the first wave of LLM-specific CA methods, including process reward models~\citep{wang2024mathshepherd,luo2024omegaprm}, token-level value estimation~\citep{vineppo}, and step-level advantage computation~\citep{gigpo}.

\paragraph{Phase 3: Agentic RL (2024--present).} The most recent phase extends RL to multi-turn, environment-interactive settings. ArCHer~\citep{archer} pioneered hierarchical multi-turn RL for LLM agents in early 2024. By 2025, agentic RL had exploded: systems trained agents for web navigation~\citep{zhou2024webarena}, software engineering (SWE-bench), scientific experimentation, and multi-agent collaboration. In this setting, trajectories span 10--100+ turns with environment interactions between each turn, total token counts reach $10^5$--$10^6$, and the reward remains sparse and terminal. The credit assignment problem is now qualitatively harder (see \Cref{sec:why_agentic_harder}), driving a second wave of innovation focused on turn-level and hindsight-based methods~\citep{hcapo,c3,ccpo,sweetrl,carl,turnppo}.

\begin{figure}[t]
    \centering
    \resizebox{\textwidth}{!}{
    \begin{tikzpicture}[
        phase/.style={draw, rounded corners=4pt, minimum height=1.2cm, text width=3.2cm, align=center, font=\small},
        p1/.style={phase, fill=gray!15},
        p2/.style={phase, fill=blue!15},
        p3/.style={phase, fill=red!15},
        arr/.style={-{Stealth[length=3mm]}, line width=1pt, gray!60},
        yr/.style={font=\scriptsize\bfseries, text=gray!70},
        desc/.style={font=\tiny, text width=3.2cm, align=center},
    ]
    \draw[gray!30, line width=0.5pt] (0,-0.6) -- (15,-0.6);

    \node[p1] (p1) at (1.8, 1.8) {Phase 1\\[2pt]\textbf{RLHF}};
    \node[desc, below=0.15cm of p1] {Dense RM reward\\Single turn, $\sim$500 tok\\CA: implicit (PPO critic)};
    \node[yr] at (1.8, -1.0) {2022--2023};
    \draw[gray!40] (1.8, -0.6) -- (1.8, 0.8);

    \node[p2] (p2) at (6.2, 1.8) {Phase 2\\[2pt]\textbf{Reasoning RL}};
    \node[desc, below=0.15cm of p2] {Outcome reward (0/1)\\Single gen, 0.5K--30K tok\\CA: token/step (PRM)};
    \node[yr] at (6.2, -1.0) {2023--2025};
    \draw[gray!40] (6.2, -0.6) -- (6.2, 0.8);

    \node[p3] (p3) at (10.6, 1.8) {Phase 3\\[2pt]\textbf{Agentic RL}};
    \node[desc, below=0.15cm of p3] {Sparse terminal reward\\Multi-turn, 100K--1M tok\\CA: turn/hindsight};
    \node[yr] at (10.6, -1.0) {2024--present};
    \draw[gray!40] (10.6, -0.6) -- (10.6, 0.8);

    \node[phase, fill=purple!10, dashed] (p4) at (14.2, 1.8) {Future\\[2pt]\textbf{Multi-Agent}};
    \node[desc, below=0.15cm of p4] {Team reward\\100+ turns\\CA: cross-agent};
    \node[yr] at (14.2, -1.0) {2026+};
    \draw[gray!40] (14.2, -0.6) -- (14.2, 0.8);

    \draw[arr] (p1) -- (p2);
    \draw[arr] (p2) -- (p3);
    \draw[arr, dashed] (p3) -- (p4);

    \node[font=\tiny\itshape, text=gray] at (7.5, -1.4) {Credit assignment difficulty $\longrightarrow$};

    \end{tikzpicture}
    }
    \caption{Evolution of RL for LLMs and the corresponding credit assignment challenges. Each phase introduces longer trajectories, more complex environments, and harder credit assignment problems. The shift from reasoning to agentic RL represents a qualitative leap in CA difficulty.}
    \label{fig:timeline}
\end{figure}

\subsection{Problem Formulation: Two MDP Abstractions}
\label{sec:formulation}

\begin{table}[t]
\centering
\caption{Summary of key notation used throughout this paper.}
\label{tab:notation}
\begin{tabular}{cl}
\toprule
\textbf{Symbol} & \textbf{Description} \\
\midrule
$x$ & Input prompt / task description \\
$y = (y_1, \ldots, y_L)$ & Generated token sequence (response) \\
$\tau$ & Complete trajectory (episode) \\
$s_t$ & State at time step $t$ \\
$a_t$ & Action at time step $t$ (token or turn-level response) \\
$o_t$ & Observation at time $t$ (in POMDP settings) \\
$R(\tau)$ & Terminal (episodic) reward for trajectory $\tau$ \\
$r_t$ & Intermediate reward at step $t$ (when available) \\
$V(s)$ & State-value function: $\mathbb{E}[R \mid s]$ \\
$Q(s, a)$ & Action-value function: $\mathbb{E}[R \mid s, a]$ \\
$\hat{A}_t$ & Estimated advantage at step $t$: $Q(s_t, a_t) - V(s_t)$ \\
$\pi_\theta$ & Policy (LLM) parameterized by $\theta$ \\
$\pi_{\text{ref}}$ & Reference policy (for DPO/KL-constrained methods) \\
$T$ & Number of turns in an agentic trajectory \\
$L$ & Sequence length (number of tokens) \\
$G$ & Group size in GRPO \\
$K$ & Number of rollouts / agents (context-dependent) \\
$\gamma$ & Discount factor \\
$\lambda$ & GAE interpolation parameter \\
$c_t$ & Credit assigned to step/turn $t$ \\
$z_u^-$ & Effective pre-unit state for an intervention unit $u$ \\
$q_u$ & Declared reference-action distribution at unit $u$ \\
$\rho$ & Downstream policy, horizon, reward/verifier, and noise protocol \\
$H(\cdot)$ & Shannon entropy \\
\bottomrule
\end{tabular}
\end{table}

\paragraph{Reasoning RL as a token-level MDP.} In reasoning RL, the model generates a single response $y = (y_1, y_2, \ldots, y_L)$ to a prompt $x$. This can be modeled as an MDP where:
\begin{itemize}[nosep]
    \item State $s_t = (x, y_1, \ldots, y_{t-1})$ is the prompt plus tokens generated so far
    \item Action $a_t = y_t$ is the next token
    \item The visible prefix transition is closed under a fixed model, tokenizer, and decoding configuration; sampled continuations can still be stochastic and hidden decoder state can matter
    \item Reward $R$ is given only at the terminal state (e.g., answer correctness)
\end{itemize}
Credit assignment here means: which tokens (or token groups) in the reasoning chain contributed to the correct answer?

\paragraph{Agentic RL as a turn-level POMDP.} In agentic RL, the model interacts with an environment over $T$ turns:
\begin{itemize}[nosep]
    \item State $s_t$ includes conversation history, environment state (partially observed), and retrieved context
    \item Action $a_t$ is the model's \emph{complete response} at turn $t$ (which itself contains many tokens)
    \item Transition is \emph{stochastic}: environment response depends on tool execution, web page state, etc.
    \item Reward $R$ is sparse and terminal (task success/failure)
\end{itemize}
Credit assignment is now \emph{doubly hierarchical}: (1) which \emph{turn} was critical? (2) within that turn, which \emph{tokens} mattered?

\paragraph{The multi-granularity action hierarchy.}
\begin{equation}
    \underbrace{\tau}_{\text{Episode}} = \underbrace{[\text{Turn}_1, \ldots, \text{Turn}_T]}_{\text{Turn level}} = \underbrace{[\text{Seg}_{1,1}, \ldots]}_{\text{Segment level}} = \underbrace{[a_{1,1,1}, \ldots]}_{\text{Token level}}
\end{equation}

\subsection{A Six-Diagnostic Lens}
\label{sec:diagnostics}

Granularity and methodology describe \emph{what a method does}; they do not by themselves explain \emph{why its evidence is admissible}. We therefore add six interacting diagnostics. A positive code means that a paper's claimed mechanism explicitly responds to the corresponding break, not that every benchmark in that setting exhibits it. In the fixed 42-paper full-text subset, the consensus positive counts are $24/10/19/20/32/7$ for transition non-closure (T), partial observability (O), limited replay (R), heterogeneous actions (H), weak local verifiability (V), and agent coupling (C), respectively. These descriptive codes are neither environment prevalence estimates nor performance comparisons.

\begin{table}[t]
\centering
\caption{Six diagnostic breaks, their identification barriers, and minimum controls. Counts refer only to the fixed 42-core-paper full-text subset.}
\label{tab:diagnostics}
\resizebox{\textwidth}{!}{
\begin{tabular}{clllr}
\toprule
\textbf{Flag} & \textbf{Diagnostic break} & \textbf{Identification barrier} & \textbf{Minimum evaluation control} & \textbf{Count} \\
\midrule
T & Transition non-closure & Same text need not imply same effective dynamics & Restore environment and hidden execution state & 24/42 \\
O & Partial observability & Logged history may omit outcome-relevant state & State sufficiency argument or privileged-state audit & 10/42 \\
R & Limited replay & Intervention cannot be reproduced exactly & Declare replay provenance and replica noise floor & 19/42 \\
H & Heterogeneous actions & Token, tool, memory, and message units are not exchangeable & Typed-unit and unit-mismatch ablations & 20/42 \\
V & Weak local verifiability & Intermediate correctness is unavailable or proxy-based & Calibrate judge/critic against intervention or continuation & 32/42 \\
C & Agent coupling & One agent's action changes another agent's policy path & Message, role, or coalition interventions & 7/42 \\
\bottomrule
\end{tabular}}
\end{table}

\paragraph{Replay fidelity is part of the estimand.} A textual prefix re-fed through a model is not equivalent to resuming the exact decode-time state (including the KV cache and sampler state), and neither operation restores an external environment checkpoint. \citet{matteson2026refeeding} shows that prefix re-feeding can introduce replica noise into token-credit estimates. Monte Carlo and counterfactual studies should therefore distinguish \emph{text re-feed}, \emph{exact decoder-state resume}, and \emph{environment checkpoint restore}; report which hidden states and random seeds are reinstated; and measure a \emph{replica noise floor} by repeating nominally identical branches. This post-audit citation feedback refines the replay diagnostic but does not change the frozen 69/92/42 denominators.

\subsection{Identification Results and the CA-ID Card}
\label{sec:identification}

Let $u$ denote a token, segment, turn, tool call, memory operation, or message edge. Let $a$ be the realized unit, $a'\sim q_u$ a declared reference, and $\rho$ the complete downstream/noise protocol. We define the protocol-specific contrast
\begin{equation}
\Delta_u(q_u;\rho)=\mathbb{E}\!\left[R(\tau^{u\leftarrow a})-R(\tau^{u\leftarrow a'})\mid z_u^-\text{ matched},\ a'\sim q_u,\rho\right].
\label{eq:ca_estimand}
\end{equation}
Different reference actions, continuation policies, horizons, verifiers, or exogenous-noise couplings define different quantities. Predictive value, semantic plausibility, information gain, and causal marginal contribution should therefore not be collapsed into an unspecified scalar called ``credit.''

\begin{insightbox}[Proposition 1: Restored-State Identification]
Suppose evaluation samples $(z_u^-,a)$ from its declared distribution and restores the same effective pre-unit state $z_u^-$ before executing $a$ and $a'\sim q_u$. If both actions have supported, consistent semantics and the two branches share the declared downstream policy, horizon, reward/verifier, and noise protocol $\rho$, then the sample mean paired return difference is unbiased for the corresponding population $\Delta_u(q_u;\rho)$. Common random numbers and independent rerolls may change variance, but not the target when their branch marginals implement the declared $\rho$.
\end{insightbox}

For logged observational comparisons, state matching is not enough: identification additionally requires conditional exchangeability $(R^a,R^{a'})\perp A_u\mid z_u^-$ and positivity for both actions~\citep{pearl2009causality,hernan2020whatif}. Critics, PRMs, judges, policy ratios, and world-model rollouts remain predictive or model-relative proxies unless calibrated against an appropriate continuation or intervention target.

\begin{challengebox}[Proposition 2: Text-Only Hidden-State Non-Identifiability]
Without restoration or valid observational assumptions, even the \emph{sign} of causal credit need not be identified from $(h_u,a_u,R)$. Let hidden $Z\sim\mathrm{Bernoulli}(1/2)$, observed text history $h$ be constant, action $A=Z$, and observed reward always equal one. Two causal worlds can agree on this entire observed distribution while setting the unobserved potential outcomes so that $\mathbb{E}[R^1-R^0]=+1/2$ in one and $-1/2$ in the other. No estimator using only text history, action, and outcome can distinguish them. Partial observability and unsupported replay are therefore identification failures, not merely noisier labels.
\end{challengebox}

These results induce four evidence tiers: \emph{restored-interventional}, \emph{observational-causal} under explicit exchangeability and positivity, \emph{model-relative counterfactual}, and \emph{predictive/proxy}. Usefulness is possible at every tier, but the claim must not exceed the evidence. \Cref{tab:ca_id_card} operationalizes this rule.

\begin{table}[t]
\centering
\caption{The CA-ID Card: a minimum claim contract, not a guarantee that causal identification is possible.}
\label{tab:ca_id_card}
\resizebox{\textwidth}{!}{
\begin{tabular}{p{2.5cm}p{5.1cm}p{5.3cm}}
\toprule
\textbf{Field} & \textbf{Required declaration} & \textbf{Evidence or falsification control} \\
\midrule
Unit and estimand & Token, segment, turn, tool, memory, or message; reference $q_u$; target contrast & Show that conclusions survive plausible unit and reference choices \\
Effective state & Contents of $z_u^-$; observed, matched, restored, or latent & Checkpoint identity or state-sufficiency argument; label text-only matching \\
Intervention provenance & Actual rerun, decoder resume, environment restore, logged match, or model/proxy & Compare weaker replay/proxy to exact or checkpointed replay on a shared subset \\
Downstream protocol & Continuation policy, horizon, reward/verifier, and noise coupling $\rho$ & Vary continuation/verifier; report common-random-number or reroll protocol \\
Support and uncertainty & Action overlap, branch repetitions, interval or power target & Wrong sign, no overlap, or failure to clear preregistered tolerance invalidates the claim \\
Optimization interface & How credit enters reward, advantage, mask, target, or loss weight & CA-specific ablation under matched model, data, optimizer, and trajectory budget \\
\bottomrule
\end{tabular}}
\end{table}

\subsection{Why GRPO's Episode-Level Credit is Insufficient}
\label{sec:grpo_limit}

The GRPO estimator~\citep{shao2024grpo} computes a group advantage:
\begin{equation}
    \hat{A}^{\text{GRPO}}_i = R(\tau_i) - \frac{1}{G}\sum_{j=1}^{G} R(\tau_j)
\end{equation}
Every token in $\tau_i$ receives the same advantage $\hat{A}^{\text{GRPO}}_i$. For a trajectory of length $L$:
\begin{itemize}[nosep]
    \item \textbf{Reasoning RL} ($L \sim 10^3$--$10^4$ tokens, 1 turn): Episode-level methods (GRPO, REINFORCE) work reasonably because the number of ``critical decisions'' is small relative to total tokens, and the signal-to-noise ratio remains manageable.
    \item \textbf{Agentic RL} ($L \sim 10^5$--$10^6$ tokens, 10--100+ turns): Episode-level methods assign identical credit to a pivotal ``choose the right API'' action and a trivial ``format the output'' action. The signal-to-noise ratio collapses.
\end{itemize}

Empirically, \citet{archer} show that standard PPO with episode-level rewards fails to learn effective multi-turn policies, while their hierarchical credit approach succeeds; \citet{ragen} report similar findings, attributing the failure to what they term the ``echo trap.''

More formally, in the REINFORCE estimator with a baseline $b$, the variance of the policy gradient for a single action $a_t$ is proportional to $(R(\tau) - b)^2$. When the same baseline is applied to all $T$ actions, the total gradient variance scales as $\mathcal{O}(T \cdot \text{Var}[R])$. GRPO and other episode-level methods mitigate this partially through group normalization, but the fundamental issue remains: with $T=100$ turns and binary reward, the signal-to-noise ratio per action is roughly $100\times$ worse than in the single-turn reasoning setting. Empirically, \citet{ragen} demonstrate this through the ``echo trap'' phenomenon: under episode-level credit, agentic models converge to repetitive behaviors because the gradient signal is too noisy to distinguish productive actions from redundant ones.

\subsection{Taxonomy Overview}
\label{sec:taxonomy}

We organize methods along two orthogonal axes (\Cref{fig:taxonomy}):

\begin{enumerate}[nosep]
    \item \textbf{Granularity axis}: At what level is credit assigned?
    \begin{itemize}[nosep]
        \item \emph{Token-level}: Individual tokens within a generation
        \item \emph{Segment-level}: Semantically meaningful spans (e.g., one reasoning step)
        \item \emph{Step/Turn-level}: A complete LLM response or tool-call cycle
        \item \emph{Multi-agent level}: Credit decomposition across collaborating agents
    \end{itemize}
    \item \textbf{Methodology axis}: How is credit computed?
    \begin{itemize}[nosep]
        \item \emph{Monte Carlo (MC)}: Rollouts from intermediate states
        \item \emph{Temporal Difference (TD)}: Learned value functions with bootstrapping
        \item \emph{Model-based / LLM-as-Critic}: LLMs evaluate intermediate states
        \item \emph{Game-theoretic}: Shapley values, counterfactual baselines
        \item \emph{Information-theoretic}: Information gain, entropy-based measures
    \end{itemize}
\end{enumerate}

\begin{figure}[t]
    \centering
    \resizebox{\textwidth}{!}{
    \begin{tikzpicture}[
        rcell/.style={fill=blue!12, rounded corners=2pt, font=\tiny, inner sep=2pt, text=blue!80!black},
        acell/.style={fill=red!12, rounded corners=2pt, font=\tiny, inner sep=2pt, text=red!80!black},
        mcell/.style={fill=purple!12, rounded corners=2pt, font=\tiny, inner sep=2pt, text=purple!80!black},
        hdr/.style={font=\footnotesize\bfseries, text=black},
    ]
    \def\cw{3.6}  
    \def\rh{2.2}  
    \def\loff{-0.3}  

    \node[hdr] at (1*\cw, 4.5*\rh) {Monte Carlo};
    \node[hdr] at (2*\cw, 4.5*\rh) {TD / Value};
    \node[hdr] at (3*\cw, 4.5*\rh) {LLM-as-Critic};
    \node[hdr] at (4*\cw, 4.5*\rh) {Game-theoretic};
    \node[hdr] at (5*\cw, 4.5*\rh) {Info-theoretic};

    \node[hdr, rotate=0, anchor=east] at (\loff*\cw, 4*\rh) {Token};
    \node[hdr, rotate=0, anchor=east] at (\loff*\cw, 3*\rh) {Segment};
    \node[hdr, rotate=0, anchor=east] at (\loff*\cw, 2*\rh) {Step/Turn};
    \node[hdr, rotate=0, anchor=east] at (\loff*\cw, 1*\rh) {Multi-Agent};

    \draw[gray!30] (0.5*\cw, 0.4*\rh) grid[xstep=\cw, ystep=\rh] (5.5*\cw, 4.4*\rh);

    \node[rcell] at (1*\cw, 4*\rh) {VinePPO};
    \node[rcell] at (2*\cw, 4.15*\rh) {RED};
    \node[rcell] at (2*\cw, 3.85*\rh) {T-REG};
    \node[rcell] at (4*\cw, 4*\rh) {From r to Q*};

    \node[rcell] at (1*\cw, 3*\rh) {SPO};
    \node[rcell] at (2*\cw, 3*\rh) {TEMPO};
    \node[rcell] at (4*\cw, 3*\rh) {SCAR};

    \node[acell] at (1*\cw, 2.2*\rh) {GiGPO};
    \node[rcell] at (1*\cw, 1.8*\rh) {SPRO};
    \node[acell] at (2*\cw, 2.25*\rh) {ArCHer};
    \node[acell] at (2*\cw, 1.95*\rh) {AgentPRM};
    \node[acell] at (2*\cw, 1.65*\rh) {SPA-RL};
    \node[rcell] at (3*\cw, 2.35*\rh) {CAPO};
    \node[acell] at (3*\cw, 2.05*\rh) {SWEET-RL};
    \node[acell] at (3*\cw, 1.75*\rh) {HCAPO};
    \node[acell] at (3*\cw, 1.45*\rh) {LaRe};
    \node[acell] at (4*\cw, 2.2*\rh) {C3};
    \node[acell] at (4*\cw, 1.9*\rh) {CCPO};
    \node[acell] at (4*\cw, 1.62*\rh) {CRAFT};
    \node[acell] at (5*\cw, 2.2*\rh) {IGPO};
    \node[acell] at (5*\cw, 1.9*\rh) {CARL};
    \node[rcell] at (1.0*\cw, 2.55*\rh) {PURE};
    \node[acell] at (2*\cw, 2.55*\rh) {iStar};
    \node[rcell] at (5*\cw, 2.55*\rh) {HICRA};
    \node[rcell] at (3.30*\cw, 2.55*\rh) {Rubric-RL};
    \node[acell] at (1.55*\cw, 1.55*\rh) {Turn-PPO};
    \node[acell] at (2.45*\cw, 1.55*\rh) {SORL};
    \node[acell] at (3.5*\cw, 1.55*\rh) {TARL};
    \node[acell] at (4.5*\cw, 1.55*\rh) {ITPO};
    \node[rcell] at (5*\cw, 1.62*\rh) {A$^2$TGPO};
    \node[acell] at (5.25*\cw, 1.42*\rh) {AEM};

    \node[mcell] at (1*\cw, 1*\rh) {M-GRPO};
    \node[mcell] at (3*\cw, 1*\rh) {LLM-MCA};
    \node[mcell] at (3.95*\cw, 1*\rh) {QLLM};
    \node[mcell] at (4.9*\cw, 1*\rh) {C3$^\dagger$};

    \node[rcell, minimum width=1.1cm] at (1.2*\cw, 0.1*\rh) {Reasoning RL};
    \node[acell, minimum width=1.1cm] at (3.0*\cw, 0.1*\rh) {Agentic RL};
    \node[mcell, minimum width=1.1cm] at (4.8*\cw, 0.1*\rh) {Multi-Agent};

    \node[font=\small\bfseries, rotate=90] at (-1.0*\cw, 2.5*\rh) {Granularity $\uparrow$};
    \node[font=\small\bfseries] at (3*\cw, -0.3*\rh) {Methodology $\rightarrow$};

    \draw[-{Stealth[length=4mm]}, line width=1.5pt, gray!50, dashed] (1.2*\cw, 3.8*\rh) -- (4.8*\cw, 1.2*\rh);
    \node[font=\scriptsize\itshape, gray!70, rotate=-35] at (3.15*\cw, 3.0*\rh) {Evolution trend};

    \end{tikzpicture}
    }
    \caption{Representative two-dimensional taxonomy of credit assignment methods for LLM RL, organized by the original assignment-granularity (vertical) and computational-methodology (horizontal) axes. \textcolor{blue!80!black}{Blue}: primarily reasoning RL; \textcolor{red!80!black}{Red}: primarily agentic RL; \textcolor{purple!80!black}{Purple}: multi-agent. The dashed arrow preserves the original reasoning-to-agentic evolution trend. The complete 69-paper taxonomy is provided in \Cref{tab:inventory}; a machine-readable frozen snapshot is planned as a dated companion-repository release.}
    \label{fig:taxonomy}
\end{figure}

\begin{figure*}[t]
\centering
\resizebox{0.97\textwidth}{!}{
\begin{forest}
forked edges,
for tree={
    grow=east,
    reversed=true,
    anchor=base west,
    parent anchor=east,
    child anchor=west,
    base=center,
    font=\normalsize,
    rectangle,
    rounded corners,
    draw,
    align=left,
    text centered,
    minimum width=4em,
    edge+={darkgray, line width=1pt},
    s sep=4pt,
    inner xsep=4pt,
    inner ysep=3pt,
    line width=0.8pt,
    ver/.style={rotate=90, child anchor=north, parent anchor=south, anchor=center},
},
where level=1{text width=6em, font=\small, fill=gray!8}{},
where level=2{text width=8em, font=\footnotesize, fill=gray!5}{},
where level=3{text width=20em, font=\scriptsize}{},
[Credit Assignment\\in LLM RL\\{\scriptsize 69 = 56C + 13E}, ver, text width=9.5em
    [Reasoning RL\\{\scriptsize 23 (21C+2E)}, fill=blue!10
        [Token-Level
            [{VinePPO (MC){,} RED (Redistrib.){,}\\T-REG (Self-gen.){,} From $r$ to $Q^*$ (Implicit){,}\\GRPO-$\lambda${,} DelTA{,} GRAIL}, fill=blue!5]
        ]
        [Segment-Level
            [{SPO (MC){,} SCAR (Shapley){,}\\TEMPO (Tree-TD){,} SCRL (Subproblem)}, fill=blue!5]
        ]
        [Step-Level
            [{PURE (PRM){,} SPRO (Masked Adv.){,}\\CAPO (LLM-Critic){,} ACPO (Attribution){,}\\HICRA (Hierarchy){,} PRL (Entropy){,} InT (Intervention){,}\\FinePO (Fine PRM){,} Rubric-RL (Judge)}, fill=blue!5]
        ]
    ]
    [Agentic RL\\{\scriptsize 40 (31C+9E)}, fill=red!10
        [Turn-Level PRM
            [{AgentPRM (TD+GAE){,} SWEET-RL (Priv.\ Critic){,}\\Turn-PPO (Turn MDP){,} SORL (Bias-corr.){,}\\TARL (LLM-Judge){,} ITPO (Implicit){,}\\Turn-Level (Hybrid)}, fill=red!5]
        ]
        [Hindsight /\\Counterfactual
            [{HCAPO (Hindsight){,} C3 (LOO){,} CCPO (Causal){,}\\CriticSearch (Retro.){,} Memory-R2 (Reroll){,}\\CRAFT (Sibling){,} PivoARL (Retry)}, fill=red!5]
        ]
        [Critic-Free /\\Step-Level
            [{GiGPO (Group-in-Group){,} POAD (Action Decomp.){,}\\CARL (Entropy){,} iStar (Implicit DPO){,} IGPO (Info.){,}\\GVPO (Verifier){,} A$^3$ (Struct.){,} A$^2$TGPO (IG/Clip)}, fill=red!5]
        ]
        [Hierarchical
            [{ArCHer (TD Hierarchy){,} PilotRL (Progressive){,}\\StepAgent (IRL){,} APPO (Branch){,} AT$^2$PO (Tree)}, fill=red!5]
        ]
        [Adjacent\\Enablers
            [{SPA-RL{,} Lightning{,} RAGEN{,} SCRIBE{,} PRS{,} AdaptSeg{,}\\AutoResearch{,} OPID{,} T$^2$PO}, fill=red!3]
        ]
    ]
    [Multi-Agent\\{\scriptsize 6 (4C+2E)}, fill=purple!10
        [Agent-Level
            [{M-GRPO (Hierarchical){,} SHARP (Shapley){,}\\MAPPA (Per-action PRM){,}\\Dr.\ MAS (Agent-wise Adv.){,}\\LLM-MCA (LLM-Critic){,} QLLM (LLM-gen.)}, fill=purple!5]
        ]
    ]
]
\end{forest}}
\caption{Hierarchical taxonomy of the reviewed credit assignment literature, preserving the original setting and family structure while adding representative frontier methods. Counts use the all-included denominator: 23 reasoning, 40 agentic or mixed, and 6 multi-agent papers, totaling 69. In parenthetical counts, C denotes a core credit-assignment method and E denotes an adjacent/boundary enabler (56C+13E overall). Leaves remain representative; \Cref{tab:inventory} is authoritative for all methods and labels, with a machine-readable frozen snapshot planned as a dated companion-repository release.}
\label{fig:taxonomy_tree}
\end{figure*}

\subsection{Classical Credit Assignment: A Brief Primer}
\label{sec:classical_ca}

Before the LLM era, deep RL developed a rich toolkit for credit assignment. We briefly review the key paradigms, as many LLM-specific methods build directly on these foundations. For a comprehensive treatment, we refer readers to \citet{pignatelli2023survey}.

\paragraph{Temporal Difference learning and value baselines.} The most widely used approach estimates a state-value function $V(s)$ and uses the advantage $A(s,a) = Q(s,a) - V(s)$ to assign credit. Generalized Advantage Estimation (GAE)~\citep{schulman2016gae} interpolates between high-bias (TD(0)) and high-variance (MC) estimates via the parameter $\lambda$:
\begin{equation}
    \hat{A}_t^{\text{GAE}(\gamma,\lambda)} = \sum_{l=0}^{\infty} (\gamma\lambda)^l \delta_{t+l}, \quad \delta_t = r_t + \gamma V(s_{t+1}) - V(s_t)
\end{equation}
In the LLM setting, AgentPRM~\citep{agentprm} directly applies TD+GAE to learn turn-level value functions for agents, while ArCHer~\citep{archer} uses an off-policy critic with TD updates.

\paragraph{Return decomposition.} RUDDER~\citep{arjona2019rudder} decomposes the episodic return into per-step contributions by training a sequence model to predict the return from partial trajectories. The contribution of step $t$ is the change in predicted return: $c_t = \hat{R}(s_{0:t}) - \hat{R}(s_{0:t-1})$. This idea directly inspires LLM methods like RED~\citep{red} (token-level redistribution), SPA-RL~\citep{sparl} (MLP-based progress estimation), and IGPO~\citep{igpo} (information gain as credit).

\paragraph{Hindsight credit assignment.} HCA~\citep{harutyunyan2019hca} reweights past actions based on the observed outcome, using the insight that knowing the future changes our estimate of which past actions were important. This ``looking backward'' principle is central to HCAPO~\citep{hcapo}, which extends hindsight credit to LLM agents using generative verification.

\paragraph{Counterfactual baselines.} Difference rewards evaluate an action's contribution by comparing the actual outcome to a counterfactual baseline: ``what would have happened if this action were replaced by a default?'' This requires either environment re-execution or model-based approximation. In the LLM setting, C3~\citep{c3} and CCPO~\citep{ccpo} implement counterfactual credit via leave-one-out analysis of agent turns, while SCAR~\citep{scar} uses Shapley values---a game-theoretic generalization of counterfactual baselines.

\paragraph{Key mapping to LLM RL.} The classical paradigms map onto LLM-specific methods as follows: TD/GAE $\to$ learned critics (ArCHer, AgentPRM); return decomposition $\to$ reward redistribution (RED, SPA-RL); hindsight $\to$ retrospective analysis (HCAPO); counterfactual $\to$ leave-one-out and Shapley (C3, SCAR). However, the LLM setting introduces a unique capability unavailable in classical RL: the \emph{LLM itself can serve as a critic}, providing natural-language evaluations of intermediate states~\citep{capo,sweetrl,lare}. This ``LLM-as-Critic'' paradigm has no direct classical analogue and represents a distinctive axis of credit assignment methodology.

\begin{insightbox}[Process Reward Models \emph{Are} Credit Assignment]
A key conceptual clarification: \textbf{Process Reward Models (PRMs)} are not merely a reward modeling technique---they are fundamentally a credit assignment mechanism. A PRM that scores each reasoning step $i$ with $r_i$ is performing step-level credit decomposition of the terminal reward $R(\tau)$. The PRM literature (Math-Shepherd, OmegaPRM, PURE) and the CA literature (VinePPO, SPRO, SCAR) are thus two perspectives on the same underlying problem. We adopt the CA perspective throughout this paper, viewing PRMs as one methodology (among several) for distributing credit.
\end{insightbox}

\subsection{RL Algorithms for LLMs: A Brief Overview}
\label{sec:rl_algorithms}

Credit assignment methods do not operate in isolation---they are components within broader RL algorithms. We briefly review the major RL algorithms used for LLM training, highlighting how each relates to credit assignment.

\paragraph{PPO (Proximal Policy Optimization).} PPO~\citep{schulman2017ppo} is the workhorse of RLHF, used in InstructGPT, ChatGPT, and Claude. PPO trains a \emph{learned value function} $V_\phi(s)$ as a baseline, computing token-level advantages via GAE. The value function is itself a credit assignment mechanism---its quality directly determines training efficiency. However, training an accurate value function for LLM-scale state spaces is notoriously difficult: the value network must process sequences of thousands of tokens and produce reliable scalar estimates, a challenge that motivates critic-free alternatives.

\paragraph{REINFORCE and REINFORCE with baseline.} The simplest policy gradient method, REINFORCE computes $\nabla_\theta J = \mathbb{E}[\sum_t \nabla_\theta \log \pi_\theta(a_t|s_t) \cdot R(\tau)]$, assigning the full return as credit to every action. Adding a baseline $b$ (e.g., the mean return) reduces variance but does not provide per-action credit differentiation. REINFORCE with learned baselines is used in several recent LLM RL systems due to its simplicity, though its credit assignment is the crudest among all approaches.

\paragraph{GRPO (Group Relative Policy Optimization).} GRPO~\citep{shao2024grpo}, introduced with DeepSeek-R1, replaces the learned value function with a \emph{group comparison baseline}: for a batch of $G$ trajectories from the same prompt, the advantage is $\hat{A}_i = R(\tau_i) - \frac{1}{G}\sum_j R(\tau_j)$. This eliminates the need for a critic network entirely, making GRPO computationally attractive. However, GRPO provides only \emph{episode-level} credit---every token in a trajectory receives the same advantage. This is the credit assignment limitation that most methods in this survey aim to improve.

\paragraph{DPO (Direct Preference Optimization).} DPO~\citep{rafailov2023dpo} bypasses explicit reward modeling by directly optimizing the policy from preference pairs. As shown by ``From $r$ to $Q^*$''~\citep{fromrToQ}, DPO implicitly learns token-level Q-values, providing an implicit form of credit assignment. Methods like iStar~\citep{istar} and ITPO~\citep{itpo} exploit this insight to extract step-level credit from DPO-trained models without explicit reward computation.

\paragraph{The credit assignment perspective on RL algorithms.} From a CA perspective, these algorithms form a spectrum: REINFORCE/GRPO provide episode-level credit (coarsest), PPO provides token-level credit via a learned critic (finer but approximate), and DPO provides implicit token-level credit (theoretically elegant but hard to extract). The methods surveyed in this paper can be viewed as \emph{enhancements} to the credit assignment quality of these base algorithms---e.g., VinePPO replaces PPO's learned critic with MC estimates, HCAPO adds hindsight analysis on top of GRPO, and CARL selectively applies credit within any base algorithm.

\paragraph{Other related algorithms.} Several other RL and self-improvement algorithms are used in LLM training but are not covered in depth because their credit assignment properties fall within the spectrum above. \emph{RLOO} (REINFORCE Leave-One-Out) uses a leave-one-out baseline $b_i = \frac{1}{G-1}\sum_{j \neq i} R(\tau_j)$, which is a variance reduction technique closely related to GRPO's group baseline; from a CA perspective, it remains episode-level. \emph{REINFORCE++} adds a token-level KL penalty to REINFORCE, occupying a middle ground between REINFORCE and PPO, but does not introduce a new credit decomposition mechanism. \emph{Online DPO, IPO, and KTO} are preference optimization variants that share DPO's implicit credit structure; their CA properties are inherited from the ``From $r$ to $Q^*$'' analysis discussed above. \emph{ReST, Expert Iteration, and STaR} are iterative self-improvement methods that filter or refine training data based on outcome quality; they interact with credit assignment indirectly (by curating which trajectories to learn from) but do not decompose credit within trajectories. We focus on PPO, GRPO, REINFORCE, and DPO because they span the main CA granularity spectrum and serve as base algorithms across the 69-paper corpus.

\section{Credit Assignment in Reasoning RL}
\label{sec:reasoning_ca}

In reasoning RL, the LLM generates a single chain-of-thought response. The trajectory is a token sequence within one generation. Credit assignment methods here operate at the \textbf{token level} and \textbf{segment/step level}, distributing the outcome reward across the reasoning chain.

\subsection{Token-Level Methods}
\label{sec:token_methods}

\subsubsection{Monte Carlo Token-Level Estimation}

\paragraph{VinePPO} \citep{vineppo}.
VinePPO (ICML 2025) replaces the learned value network in PPO with Monte Carlo continuation estimates at the token level. At position $t$, it samples $K$ continuations from an intermediate prefix, evaluates them under the declared outcome reward, and estimates $V_{\rho}(h_t) \approx \frac{1}{K}\sum_{k=1}^{K} R(\tau_t^{(k)})$ for that continuation/re-feed protocol $\rho$. The token-level advantage is then $\hat{A}_t = R(\tau)-V_{\rho}(h_t)$. This avoids learned-critic approximation but is not automatically an intervention effect or an exact-state value: text re-feeding can differ from resuming the original decoder/KV state, and the estimator is unbiased only for the declared sampling protocol when its i.i.d. and reward assumptions hold~\citep{matteson2026refeeding}. The paper reports improvements over its within-study PPO baselines on GSM8K and MATH. Computational cost scales as $\mathcal{O}(K\cdot L)$ additional continuations, and replay fidelity should be reported with a replica noise floor (\Cref{sec:diagnostics}).

\subsubsection{Reward Redistribution}

\paragraph{RED} \citep{red}.
RED (Reward Redistribution to Token Level) takes a pragmatic approach: given an off-the-shelf reward model (RM) trained for RLHF, it probes the RM's internal representations to estimate token-level reward contributions via linear regression. Specifically, it trains a lightweight probe on the RM's hidden states to predict each token's marginal contribution to the overall reward score. This requires zero additional RL training---the redistribution is purely post-hoc. Despite its simplicity, RED provides a surprisingly effective token-level signal that improves PPO training over uniform credit assignment, suggesting that pre-trained reward models already encode rich credit assignment information that is underutilized.

\paragraph{T-REG} \citep{treg}.
T-REG (Token-Level Reward Regularization) generates token-level reward signals without any external model. It uses a contrastive self-prompting strategy: for a given problem, the model generates both correct and incorrect solutions, then compares the token-level log-probability differences to identify which tokens are most discriminative. Tokens that differ most between correct and incorrect solutions receive higher credit. This self-supervised approach is elegant in its simplicity and requires no reward model, critic, or additional rollouts.

\subsubsection{Implicit Token-Level Credit}

\paragraph{From $r$ to $Q^*$} \citep{fromrToQ}.
This work provides a theoretical foundation for implicit credit assignment in preference-trained models. It shows that DPO (Direct Preference Optimization) implicitly learns a token-level Q-function: the log-probability ratio between the trained and reference models at each token position corresponds to a soft Q-value under the Bellman equation. Formally, $Q^*(s_t, a_t) = \beta \log \frac{\pi_\theta(a_t|s_t)}{\pi_{\text{ref}}(a_t|s_t)} + \beta \log Z(s_t)$, where $\beta$ is the DPO temperature and $Z$ is a normalizing partition function. This insight implies that any preference-trained LLM already encodes credit assignment information, and extracting this implicit credit is potentially more efficient than learning explicit reward models. The practical implication is profound: credit assignment may be a ``free'' byproduct of alignment training.

\subsection{Segment-Level Methods}
\label{sec:segment_methods}

\paragraph{SPO} \citep{spo}.
SPO (Segment Policy Optimization) identifies a practical middle ground between token-level and episode-level credit. It partitions the reasoning chain into semantically meaningful \emph{segments} at ``cutpoints''---positions where the reasoning transitions between distinct sub-problems or approaches (e.g., between setting up an equation and solving it). For each segment, SPO computes an MC advantage by comparing the outcomes of trajectories that share the same prefix up to that segment. This segment-level granularity naturally aligns with the structure of mathematical reasoning, where each ``step'' is a coherent unit, while avoiding the prohibitive cost of token-level MC estimation.

\paragraph{TEMPO} \citep{tempo}.
TEMPO (Tree-Structured Credit Assignment) generalizes the linear chain structure of reasoning to a tree. At decision points where the model could have taken different paths, TEMPO branches the trajectory into alternative continuations and applies \emph{branch-gated TD corrections}: sampled leaf returns are propagated upward with TD-style bootstrapping. The leaf averages estimate the declared branch protocol, while internal bootstrapping trades additional approximation for variance control. TEMPO is \emph{critic-free} in the sense that it does not train a separate learned value function; its validity still depends on branch provenance and replay fidelity.

\paragraph{SCAR} \citep{scar}.
SCAR (Shapley Credit Assignment Rewards) brings cooperative game theory to credit assignment. It treats the reasoning chain as a coalitional game where each segment is a ``player,'' and the outcome reward is the game's value. Each segment's credit is its \emph{Shapley value}---the average marginal contribution across all possible orderings of segments. The Shapley value is the unique attribution method satisfying efficiency (credits sum to total reward), symmetry (equal contributors receive equal credit), and the null player property (non-contributors receive zero credit). The main challenge is computational: exact Shapley values require evaluating $2^n$ coalitions for $n$ segments. SCAR uses sampling-based approximations, trading exactness for tractability. Despite the overhead, SCAR provides a theoretically principled credit assignment that can serve as a gold-standard reference for evaluating cheaper heuristic methods.

\subsection{Step-Level Methods in Reasoning}
\label{sec:reasoning_step}

These methods treat each ``reasoning step'' (e.g., one line of math derivation) as the unit of credit.

\subsubsection{Process Reward Models (PRMs)}

\paragraph{Background: Math-Shepherd and OmegaPRM.}
The Process Reward Model (PRM) paradigm, introduced for reasoning verification, provides a natural framework for step-level credit assignment. Math-Shepherd~\citep{wang2024mathshepherd} pioneered automatic step-level labeling: for each reasoning step, it samples multiple continuations and labels the step as ``correct'' if a sufficient fraction of continuations reach the right answer. OmegaPRM~\citep{luo2024omegaprm} scaled this approach using a divide-and-conquer strategy that efficiently explores the tree of possible continuations. These PRM foundations provide the step-level supervision that downstream CA methods build upon, and their MC-based labeling strategy directly connects to the classical return decomposition paradigm.

\paragraph{PURE} \citep{pure}.
PURE (ICML 2025) makes a subtle but important theoretical contribution to PRM-based credit. Standard PRMs assign step-level value as the \emph{expected sum} of future rewards: $V(s_t) = \mathbb{E}[\sum_{t'=t}^{T} r_{t'}]$. PURE argues this ``sum-form'' credit is vulnerable to reward hacking---models can learn to produce ``safe'' intermediate steps that inflate the expected sum without actually contributing to correctness. Instead, PURE proposes \emph{min-form} credit: $V(s_t) = \mathbb{E}[\min_{t' \geq t} r_{t'}]$, where the value of a state is determined by the \emph{worst} future step. This prevents the model from ``hiding'' errors behind high-scoring steps and provides more robust step-level credit signals. The theoretical analysis shows that min-form credit leads to better-calibrated process rewards and reduced overoptimization.

\paragraph{SPRO} \citep{spro}.
SPRO (Self-Guided Process Reward) introduces a self-supervised approach to step-level credit that requires no external PRM or reward model. Its core mechanism is the \emph{Masked Step Advantage}: for each step $i$ in a solution, SPRO masks (removes) the step and re-evaluates the solution's likelihood of reaching the correct answer. The credit for step $i$ is the performance drop caused by its removal: $c_i = P(\text{correct} | \text{full solution}) - P(\text{correct} | \text{solution without step } i)$. This leave-one-out approach provides an intuitive measure of each step's necessity. SPRO reports a $3.4\times$ improvement in training efficiency over standard GRPO, demonstrating that even simple self-supervised credit signals can dramatically accelerate learning.

\paragraph{FinePO} \citep{fineprm}.
FinePO (part of the SketchVL framework for chart understanding) demonstrates that the PRM paradigm can be pushed to \emph{sub-step} granularity in domain-specific settings. Within a visual reasoning pipeline, FinePO scores individual operations within each reasoning step, providing finer credit signals than standard step-level PRMs. While developed for a specific domain (chart and diagram understanding rather than general mathematical reasoning), its credit assignment mechanism---decomposing step-level rewards into sub-step contributions---illustrates a direction that may generalize to other settings where reasoning steps have internal structure.

\paragraph{PRL} \citep{prl}.
PRL (Process Reward Learning) provides a theoretically elegant connection between process rewards and the structure of optimal policies. It derives step-level process rewards from the decomposition of entropy-regularized RL objectives, showing that the optimal process reward at each step equals the advantage function under the entropy-regularized optimal policy. This theoretical grounding means PRL's credit signals are not heuristic but provably optimal under specific assumptions, providing a principled foundation for step-level credit assignment.

\paragraph{InT} \citep{int}.
InT (Self-Proposed Interventions) takes a unique approach to credit assignment in reasoning: the model itself proposes \emph{interventions}---counterfactual modifications to specific reasoning steps---and evaluates whether these interventions change the outcome. Steps where interventions alter the result receive high credit; steps where interventions are inconsequential receive low credit. This self-proposed intervention mechanism provides a principled, model-intrinsic measure of step importance without external reward models.

\subsubsection{Attribution-Based and Curriculum Methods}

\paragraph{ACPO} \citep{acpo}.
ACPO (Attribution-based Credit for RLVR) combines credit assignment with curriculum learning. It computes factorized hierarchical rewards that decompose the outcome reward into step contributions using attribution methods (e.g., gradient-based saliency), then uses these step-level signals to construct a difficulty-aware training curriculum. Problems where credit is concentrated on a few steps (clear bifurcation points) are prioritized early in training, while problems with diffuse credit (many steps contribute equally) are introduced later. This synergy between credit assignment and data selection exemplifies a broader trend: CA is not just about reward redistribution but about making the entire training pipeline more efficient.

\subsubsection{LLM-as-Critic for Reasoning}

\paragraph{CAPO} \citep{capo}.
CAPO (Credit Assignment Policy Optimization) exploits a capability unique to the LLM setting: the model can serve as its own critic. CAPO uses the LLM as a \emph{Generative PRM} (GenPRM)---given a reasoning trajectory, the same LLM (or a prompted version of it) generates natural-language critiques of each step, assessing correctness, relevance, and contribution to the final answer. These critiques are converted into scalar step-level rewards that drive policy optimization. The key advantage is self-containment: no separate reward model, critic network, or MC rollouts are needed. The key risk is \emph{self-evaluation bias}---the model may systematically overrate its own steps---which CAPO mitigates through calibration techniques.

\subsubsection{Hierarchy-Aware Methods in Reasoning}

\paragraph{HICRA} \citep{hicra}.
HICRA (Hierarchy-Aware Credit Assignment) studies how RL develops hierarchical reasoning in LLMs. It identifies a two-phase learning dynamic: models first acquire \emph{procedural skills} (routine computations) and then develop \emph{strategic planning} (high-level problem decomposition). HICRA proposes focusing credit on high-impact planning tokens rather than distributing learning signals uniformly, showing that this hierarchy-aware approach significantly outperforms flat credit assignment. While HICRA is developed in the reasoning RL context, its insight---that different \emph{functional roles} of tokens (planning vs.\ procedural) deserve different credit treatment---is highly relevant to agentic settings (see \Cref{sec:agentic_hierarchical}), where the distinction between strategic decisions and routine execution is even more pronounced.

\subsection{Discussion: The State of Credit Assignment in Reasoning RL}
\label{sec:reasoning_discussion}

The methods reviewed in this section reveal a maturing landscape with clear trade-offs:

\begin{itemize}[nosep]
    \item \textbf{Token-level methods} (VinePPO, RED, T-REG) provide the finest credit granularity but face computational challenges. VinePPO's MC approach is theoretically principled but expensive; RED and T-REG offer cheaper alternatives at the cost of less rigorous credit signals.
    \item \textbf{Segment/step-level methods} represent the current mainstream, with PRMs (PURE, SPRO) and hierarchy-aware approaches (HICRA) offering practical balances between credit quality and computational cost. Domain-specific extensions like FinePO~\citep{fineprm} demonstrate that sub-step granularity is feasible in structured domains.
    \item \textbf{The LLM-as-Critic paradigm} (CAPO) is emerging as a distinctive LLM-native approach that has no classical RL analogue.
\end{itemize}

A critical observation is that many reasoning RL credit assignment methods benefit from three conditions, none of which follows merely from using text:
\begin{enumerate}[nosep]
    \item \textbf{Prefix/effective-state closure}: the recorded prefix and declared decoder state are sufficient for the continuation target being estimated.
    \item \textbf{Branchability and checkpointability}: continuations can be generated from a controlled prefix or restored state with known replay provenance and measurable replica noise.
    \item \textbf{Verifiable outcomes}: the final answer, and sometimes intermediate steps, can be checked under a declared verifier.
\end{enumerate}

When these conditions fail, the corresponding estimand or evaluation must change. Extending a vine estimator to an agentic turn, for example, requires a specified environment restore or model-relative branch protocol; PRMs require a locally meaningful verifier or must be labeled as proxies.

The success of credit assignment in reasoning RL raises a natural question: \emph{can the same methods work when LLMs interact with real environments?} As we characterize in the next section, the answer is largely no---agentic RL introduces qualitatively new challenges that call for different approaches.

\section{Why Agentic RL Fundamentally Reshapes Credit Assignment}
\label{sec:why_agentic_harder}

Before reviewing agentic CA methods, we characterize \emph{what makes credit assignment in agentic RL qualitatively different} from reasoning RL. This section provides the conceptual foundation for understanding why new methods are needed.

\subsection{Challenge 1: Transition Non-Closure and Replay}
\label{sec:challenge_stochastic}

In many reasoning experiments, a fixed model and textual prefix provide a convenient branch point for sampling continuations. This is \emph{prefix closure for a declared continuation protocol}, not a guarantee that re-feeding text restores the original decode-time/KV state. Vine-style estimates are therefore continuation values relative to that protocol; exact-state claims require decoder-state resume or a measured replica noise floor~\citep{vineppo,matteson2026refeeding}.

Agentic interaction more often breaks effective-state closure. After a tool call, web request, or code command, the next observation may depend on hidden or mutable environment state:
\begin{itemize}[nosep]
    \item API calls may fail, timeout, or return rate-limited responses.
    \item Web pages may have changed since the last access, or load differently due to A/B testing.
    \item Code execution may produce non-deterministic outputs (e.g., floating-point variations, race conditions).
    \item In conversational settings, user responses are inherently unpredictable.
\end{itemize}

These failures affect both estimation and claim scope. MC methods need an environment checkpoint, logged-support argument, or explicitly model-relative rerun; TD methods face hidden-state bias in addition to transition variance. Retrospective methods~\citep{hcapo} avoid some real re-execution cost, but an LLM reconstruction remains a proxy unless calibrated against restored branches.

\subsection{Challenge 2: Partial Observability}
\label{sec:challenge_pomdp}

Reasoning studies often treat prompt plus prefix as a sufficient observed state, although hidden decoder state can violate that abstraction. Agentic RL more visibly exposes partial observability: the agent receives a textual observation $o_t=\mathcal{O}(s_t)$ that can omit outcome-relevant environment state:
\begin{itemize}[nosep]
    \item The full state of a database is not visible---the agent sees only query results.
    \item File system contents are observed only through explicit \texttt{ls} or \texttt{cat} commands.
    \item In multi-agent settings, other agents' internal states and reasoning are hidden.
    \item Web page state includes invisible elements (JavaScript state, session data, server-side logic).
\end{itemize}

Partial observability fundamentally complicates credit assignment because it introduces \emph{ambiguity between decision quality and information availability}. An action that appears ``bad'' in hindsight (e.g., calling the wrong API) may have been \emph{optimal given the agent's information} at the time. A correct credit assignment system must distinguish between:
\begin{enumerate}[nosep]
    \item \textbf{Decision errors}: the agent had sufficient information but chose poorly.
    \item \textbf{Information gaps}: the agent lacked critical information and no available action could have bridged the gap.
    \item \textbf{Exploratory actions}: the agent correctly chose to gather information, even if the immediate outcome was negative.
\end{enumerate}
Most current CA methods do not explicitly address this distinction, assigning credit based on outcomes rather than decision quality relative to available information. Addressing this gap is an important open problem (see \Cref{sec:open}).

\subsection{Challenge 3: Vastly Longer Horizons}
\label{sec:challenge_horizon}

The quantitative difference in trajectory length between reasoning and agentic RL is dramatic:

\begin{table}[t]
\centering
\caption{Trajectory complexity across reasoning and agentic RL settings. Agentic tasks involve dramatically more turns, tokens, and decision points, creating qualitative challenges for credit assignment.}
\label{tab:horizon_comparison}
\begin{tabular}{lccc}
\toprule
\textbf{Setting} & \textbf{Turns} & \textbf{Tokens} & \textbf{``Decision points''} \\
\midrule
Reasoning RL (GSM8K) & 1 & 200--800 & 3--10 steps \\
Reasoning RL (MATH) & 1 & 1,000--5,000 & 5--20 steps \\
Reasoning RL (AIME/competition) & 1 & 10,000--30,000+ & 20--100 steps \\
Agentic RL (ALFWorld/WebShop) & 5--20 & 5,000--30,000 & 5--20 turns \\
Agentic RL (WebArena) & 10--30 & 30,000--100,000 & 10--30 turns \\
Agentic RL (SWE-bench) & 20--100+ & 100,000--500,000+ & 20--100+ turns \\
Agentic RL (OSWorld) & 50--100 & 200,000--1,000,000 & 50--100+ turns \\
\bottomrule
\end{tabular}
\end{table}

This is not merely a quantitative difference: broadcasting one terminal scalar over more decisions increases the number of jointly updated actions without revealing which local choice mattered. The resulting policy-gradient variance depends on action-score covariance, the baseline, and the trajectory distribution; it does not obey a universal horizon-only multiplier. In practice, RAGEN reports an ``echo trap'' in which agents converge to repetitive safe behaviors under sparse episode feedback~\citep{ragen}. This is within-study evidence of a failure mode, not a general scaling law.

Moreover, long horizons create a \emph{temporal distance} problem: early decisions (e.g., choosing a problem-solving strategy at turn 1) have consequences that only manifest many turns later. The causal chain between action and outcome becomes increasingly indirect, making both MC and TD approaches less effective.

\subsection{Challenge 4: Heterogeneous Action Types}
\label{sec:challenge_heterogeneous}

In reasoning RL, actions are homogeneous: every action is ``generate the next token'' or ``produce the next reasoning step.'' The credit profile of actions is relatively uniform---each step contributes incrementally to the solution.

Agentic RL introduces radical \emph{action heterogeneity}. Within a single trajectory, an agent may perform:
\begin{itemize}[nosep]
    \item \textbf{Planning actions}: formulating a high-level strategy (``I should first search for the API documentation, then write a test, then implement the function'').
    \item \textbf{Tool selection}: choosing \emph{which} tool to invoke (search vs. calculator vs. code execution).
    \item \textbf{Tool parameterization}: deciding \emph{how} to invoke the tool (what query to search, what code to run).
    \item \textbf{Communication}: sending messages to users or other agents.
    \item \textbf{Error recovery}: detecting failures and deciding how to retry or pivot.
    \item \textbf{Bookkeeping}: formatting outputs, updating internal state, logging progress.
\end{itemize}

These action types have vastly different ``credit profiles.'' A wrong tool selection at a critical juncture can be catastrophic (leading to a completely wrong solution path), while a suboptimal output format is trivial. Episode-level credit assigns equal weight to both. This heterogeneity motivates methods like CARL~\citep{carl}, which uses action entropy to identify high-impact decision points and focuses credit there, and HICRA~\citep{hicra}, which distinguishes between ``planning tokens'' and ``procedural tokens'' in the reasoning setting.

\subsection{Challenge 5: Non-Verifiable Intermediate States}
\label{sec:challenge_verifiability}

A crucial enabler for credit assignment in reasoning RL is \emph{step-level verifiability}. In mathematical reasoning, each intermediate step can often be checked: ``Is this algebraic manipulation correct?'' ``Does this equation follow from the previous one?'' This verifiability underpins the entire process reward model (PRM) paradigm~\citep{wang2024mathshepherd,luo2024omegaprm,pure}, where step-level labels ($+$/$-$) provide dense supervision for credit assignment.

In agentic RL, intermediate verification is rarely possible:
\begin{itemize}[nosep]
    \item \textbf{Tool calls}: Is ``\texttt{search(`Python web scraping')}'' a good action? It depends entirely on what the search returns, which is unknown until after execution.
    \item \textbf{Code generation}: Is the generated code correct? Only verifiable after execution, and even then, partial correctness is hard to quantify.
    \item \textbf{Navigation}: Is clicking on link X productive? Depends on where it leads.
    \item \textbf{Communication}: Is ``asking the user for clarification'' helpful? Subjective and context-dependent.
\end{itemize}

The absence of intermediate verifiability means that PRM-style methods, which are the most established approach in reasoning RL, cannot be directly transferred to agentic settings. This gap drives the development of alternative approaches: hindsight-based credit~\citep{hcapo} (evaluate actions \emph{after} seeing outcomes), implicit credit via DPO~\citep{istar} (avoid explicit step-level evaluation entirely), and privileged critics~\citep{sweetrl} (use information available only at training time to provide step-level signals).

\subsection{Challenge 6: The Bifurcation Point Problem}
\label{sec:challenge_bifurcation}

We define a \textbf{bifurcation point} as a state where the agent's action has an outsized impact on the trajectory outcome---a ``fork in the road'' where different choices lead to dramatically different results. In agentic RL, bifurcation points have distinctive characteristics:

\begin{itemize}[nosep]
    \item \textbf{Rarity}: Most actions in an agentic trajectory are ``routine''---following obvious next steps, formatting outputs, making standard API calls. Empirical analysis in CARL~\citep{carl} suggests that bifurcation points may occur at only a small fraction of decision points.
    \item \textbf{Decisiveness}: Despite their rarity, bifurcation points can account for a disproportionate share of outcome variance. Choosing the right debugging strategy, selecting the correct tool for a task, or formulating an effective search query are often the actions that separate success from failure.
    \item \textbf{Non-obviousness}: Bifurcation points are often not identifiable in advance---their importance only becomes clear in retrospect, when we can see how the trajectory unfolded.
\end{itemize}

Episode-level credit (GRPO) is blind to bifurcation points: it assigns equal credit to a pivotal tool selection and a trivial formatting action. This motivates two complementary strategies: (1) \emph{identify} bifurcation points and focus credit there (CARL~\citep{carl} uses action entropy as a proxy; HICRA~\citep{hicra} distinguishes planning from procedural actions), and (2) \emph{evaluate} bifurcation points retrospectively (HCAPO~\citep{hcapo} uses hindsight analysis; C3~\citep{c3} uses counterfactual comparison).

\subsection{Summary: The Agentic Credit Assignment Gap}

\begin{table}[t]
\centering
\caption{Typical diagnostic conditions in reasoning and agentic studies; these are tendencies, not defining binaries.}
\label{tab:reasoning_vs_agentic}
\resizebox{\textwidth}{!}{
\begin{tabular}{lcc}
\toprule
\textbf{Dimension} & \textbf{Reasoning RL} & \textbf{Agentic RL} \\
\midrule
Effective-state closure & Often assumed from prefix & Often broken by hidden environment state \\
Replay provenance & Text branch or decoder resume & Environment checkpoint, logged match, or proxy \\
Observability & Prefix often treated as sufficient & Partial environment observation common \\
Typical horizon & 1 turn, 0.5K--30K tokens & 10--100+ turns, 100K--1M tokens \\
Action types & Homogeneous (tokens) & Heterogeneous (tool, plan, communicate) \\
Intermediate verification & Often possible (math) & Rarely possible \\
Bifurcation points & Moderate frequency & Rare but decisive \\
\bottomrule
\end{tabular}
}
\end{table}

\section{Credit Assignment in Agentic RL}
\label{sec:agentic_ca}

We now review methods specifically designed for or applicable to agentic RL, where multi-turn environment interaction is central.

\subsection{Turn-Level Process Reward Models}
\label{sec:agentic_prm}

\paragraph{AgentPRM} \citep{agentprm}.
AgentPRM adapts the PRM paradigm from reasoning to agentic settings by replacing MC-based step labeling with \emph{TD+GAE-based value estimation}. The key insight is that MC labeling---sampling continuations from each step to estimate step correctness---is prohibitively expensive in agentic settings because it requires re-executing environment interactions (spinning up sandboxed environments, making real API calls, etc.). Instead, AgentPRM trains a step-level critic using temporal difference learning: $V(s_t) \leftarrow V(s_t) + \alpha[r_t + \gamma V(s_{t+1}) - V(s_t)]$, with GAE for advantage estimation. Applied to tool-use, code generation, and web navigation tasks, AgentPRM reports $8\times$ better sample efficiency compared to MC-based PRM training. This work demonstrates that the TD paradigm, while introducing bias through bootstrapping, is practically necessary when environment re-execution is costly.

\paragraph{SWEET-RL} \citep{sweetrl}.
SWEET-RL (Meta/FAIR) introduces the concept of a \emph{privileged (asymmetric) critic} for multi-turn LLM agent training. The core idea exploits the training/inference asymmetry: at training time, we have access to information that the agent does not have at inference time---specifically, the ground truth answer, the complete future trajectory, and possibly environment state variables. SWEET-RL trains a critic that conditions on this privileged information to provide high-quality turn-level reward signals, which are then used for DPO-style optimization of the actor (which sees only the standard observation). This approach elegantly sidesteps the non-verifiability challenge (\Cref{sec:challenge_verifiability}): even when intermediate states cannot be verified from the agent's perspective, the privileged critic can evaluate them using information available only during training. The asymmetric design ensures that the actor's policy is optimized for the actual (partially observable) setting, while the credit signals benefit from the full information available during training.

\paragraph{Turn-Level Reward Design} \citep{turnlevel}.
This work (NeurIPS 2025) proposes a hybrid reward design that matches the reward type to the action type. For turns whose outputs are \emph{verifiable} (e.g., code execution results, database query outputs, mathematical calculations), it uses automated verification to provide exact turn-level rewards. For turns whose outputs are \emph{subjective} or hard to verify (e.g., planning, information synthesis, communication), it employs an LLM-as-judge to provide approximate turn-level scores. The framework formalizes multi-turn agent training as an MDP with heterogeneous reward sources and shows that this hybrid approach significantly outperforms both pure verification-based and pure LLM-judge-based rewards, as each reward type is applied where it is most reliable.

\paragraph{Turn-PPO} \citep{turnppo}.
Turn-PPO (EACL 2026) reformulates multi-turn agent RL as a \emph{turn-level MDP}, where each complete response plus environment feedback is treated as a macro-action. It computes turn-level advantages and importance ratios rather than broadcasting a turn signal through a token-level ratio without adjustment. The paper reports improved stability and final performance over its within-study PPO baselines on WebShop and Sokoban, supporting turns as a useful optimization unit in those evaluated settings rather than establishing a universal atomic unit.

\paragraph{SORL} \citep{stppo}.
SORL (Stabilizing Off-Policy RL for Long-Horizon Agent Training) addresses instability in multi-turn agent RL caused by two sources: (1) the granularity mismatch between token-level optimization and turn-structured interactions, and (2) high-variance gradient updates from off-policy sampling. SORL proposes turn-level importance sampling combined with clipping-triggered normalization, instantiated as two algorithms---SO-PPO and SO-GRPO---that align policy optimization with the structure of multi-turn interactions and adaptively suppress unreliable off-policy updates. Evaluated on multi-turn search benchmarks, SORL provides theoretical grounding for why turn-level CA requires purpose-built optimization algorithms rather than naive application of standard PPO or GRPO.

\paragraph{TARL} \citep{tarl}.
TARL (Turn-level Adjudicated Reinforcement Learning) proposes a process-supervised RL framework for interactive multimodal tool-use agents. Its core mechanism employs an LLM as a judge to provide turn-level evaluation during training, addressing the credit assignment challenge in long-horizon agentic tasks. Combined with a mixed-task training curriculum that integrates mathematical reasoning problems, TARL reports a 6\%+ improvement in task pass rate on the $\tau$-bench benchmark over strong RL baselines, demonstrating the value of turn-level process supervision for multi-modal agents.

\paragraph{ITPO} \citep{itpo}.
ITPO (Implicit Turn-Level Process Rewards, March 2026) derives implicit turn-level process rewards from sparse outcome signals without training a separate reward model. Building on the ``From $r$ to $Q^*$'' insight~\citep{fromrToQ}, ITPO extracts turn-level rewards from the model's own log-probability changes across turns, treating the policy itself as an implicit critic. Applied to proactive multi-turn interaction settings (tutoring, recommendation), ITPO shows that implicit turn-level credit is competitive with explicitly trained turn-level critics at a fraction of the computational cost.

\subsection{Hindsight and Counterfactual Methods}
\label{sec:agentic_hindsight}

These methods exploit a key advantage of post-hoc analysis: after the trajectory is complete, we can look back and reason about what mattered.

\paragraph{HCAPO} \citep{hcapo}.
HCAPO (Hindsight Credit Assignment for Policy Optimization, March 2026) addresses weak intermediate verifiability through retrospective analysis. After collection, an LLM critic conditions on the completed trajectory and generates alternative continuations for each turn. This can provide a richer semantic training signal without real environment re-execution, but it estimates a \emph{model-relative} counterfactual: knowing the outcome does not by itself distinguish luck from causal contribution. Stronger causal claims require calibration against restored environment branches or justified observational assumptions (\Cref{sec:identification}).

\paragraph{C3} \citep{c3}.
C3 (Contextual Counterfactual Credit Assignment, March 2026) uses a leave-one-out framework: $c_t=R(\tau)-\widehat{R}(\tau_{\setminus t})$, where a declared default replaces turn $t$. Because the counterfactual is generally not re-executed in the environment, an LLM approximates how the trajectory might have unfolded. The resulting score is a semantic/model-relative marginal contribution unless calibrated against restored interventions; changing the default or continuation model changes the estimand. The framework can be applied to agent or turn units, subject to the same qualification.

\paragraph{CCPO} \citep{ccpo}.
CCPO (Counterfactual Credit Policy Optimization, March 2026) casts each turn action as a treatment in a structural causal model and targets a turn-level effect under stated assumptions. Its formal result is conditional on the specified graph, intervention semantics, support, and absence of relevant unobserved confounding. A full conversation history does not establish those conditions when decoder, tool, user, or environment state is hidden. Practical model-based substitutions should therefore be labeled model-relative or proxy estimates unless calibrated against restoration; the contemporaneous HCAPO, C3, and CCPO papers show active interest in retrospective credit, not consensus that one paradigm is universally identified.

\paragraph{CriticSearch} \citep{criticsearch}.
CriticSearch applies retrospective credit assignment specifically to \emph{search agents}---LLMs that issue search queries, process results, and iteratively refine their answers. A frozen, asymmetric critique LLM retrospectively evaluates each search turn using privileged information (the full trajectory and gold answers), converting these assessments into dense, turn-level rewards. This is closely related to SWEET-RL's privileged critic design (\Cref{sec:agentic_prm}), but specialized for the search domain where each turn involves a distinct query-result cycle. CriticSearch reports improved convergence speed and stability on multi-hop reasoning benchmarks, demonstrating that retrospective critics are effective even in information-retrieval-centric agent tasks.

\subsection{Critic-Free Step-Level Methods}
\label{sec:agentic_criticfree}

\paragraph{GiGPO} \citep{gigpo}.
GiGPO (Group-in-Group Policy Optimization, NeurIPS 2025) extends GRPO's group comparison principle from episodes to steps. Its outer level compares trajectories; its inner level groups steps by anchor-state identifiers and computes within-group relative advantages, without a separately trained critic. The paper reports gains over its matched GRPO baselines on ALFWorld and WebShop. Those within-study results support the proposed grouping under the evaluated state-matching procedure, but do not by themselves identify causal step contribution or generalize across environments.

\paragraph{POAD} \citep{poad}.
POAD (Policy Optimization with Action Decomposition) addresses a subtle issue in agentic RL: the discrepancy between action-level and token-level optimization. In agentic settings, each ``action'' (e.g., a tool call or response) is a variable-length token sequence, yet standard RL treats it as atomic. POAD derives \emph{Bellman backup with Action Decomposition} (BAD), which integrates credit assignment at two levels: \emph{intra-action} (distributing credit across tokens within a single action) and \emph{inter-action} (distributing credit across sequential actions). This decomposition is implemented within PPO, yielding enhanced learning efficiency and generalization. POAD is notable as one of the earliest (May 2024) methods to formalize the action-to-token credit decomposition problem for LLM agents.

\subsection{Hierarchical Methods}
\label{sec:agentic_hierarchical}

Agentic tasks have natural hierarchies (plan → execute → verify). These methods exploit this structure.

\paragraph{ArCHer} \citep{archer}.
ArCHer (ICML 2024) is the pioneering work on hierarchical credit assignment for multi-turn LLM agents. It introduces an explicit two-level architecture: a \emph{high-level off-policy critic} that learns a turn-level Q-function $Q^H(s_t, a_t)$ (where $a_t$ is the complete LLM response at turn $t$), and a \emph{low-level on-policy actor} that optimizes the token-level policy $\pi_\theta(y|s_t)$ within each turn. The high-level critic is trained with off-policy TD updates, enabling sample-efficient learning from a replay buffer of past trajectories. The low-level actor is optimized on-policy using the high-level Q-values as turn-level rewards. This decoupled architecture directly addresses the doubly-hierarchical credit assignment challenge: the high-level critic handles \emph{which turn matters}, while the low-level actor handles \emph{which tokens within that turn matter}. ArCHer was the first to formally recognize that multi-turn LLM RL requires fundamentally different credit assignment than single-turn reasoning RL.

We note that HICRA~\citep{hicra}, reviewed in \Cref{sec:reasoning_ca}, provides a reasoning-RL foundation for hierarchy-aware credit that directly informs the agentic methods in this section. Its distinction between planning and procedural tokens maps naturally to the plan-execute hierarchy of agentic tasks.

\paragraph{PilotRL} \citep{pilotrl}.
PilotRL (Global Planning-Guided Progressive RL) extends the hierarchical principle to a three-stage progressive framework: (1) \emph{plan-level RL}, where credit is assigned to high-level plan components; (2) \emph{step-level RL}, where credit is refined within each plan component; (3) \emph{token-level RL}, where credit cascades to individual tokens. Credit flows from coarse to fine across stages, with each stage providing the reward signal for the next. This cascaded approach is designed for agents that explicitly formulate plans before executing them (e.g., ``step 1: search for relevant files; step 2: understand the codebase; step 3: implement the fix'').

\paragraph{CARL} \citep{carl}.
CARL (NeurIPS 2025) addresses heterogeneous actions by using policy entropy as a proxy for selecting \textbf{critical actions} and applying updates to a subset of decision points. In the paper's evaluated tasks and training recipe, this selection used 72\% fewer gradient updates while matching the reported full-update baseline. The result establishes a within-study efficiency finding; entropy is not itself causal contribution, and it does not imply that the remaining actions generally have negligible credit.

\subsection{Information-Theoretic Methods}
\label{sec:agentic_info}

\paragraph{IGPO} \citep{igpo}.
IGPO (Information Gain Policy Optimization) takes an information-theoretic approach to turn-level credit. For each turn $t$, IGPO defines the credit as the \emph{information gain} about task success:
\begin{equation}
    c_t = \log P(\text{success} | h_{1:t}) - \log P(\text{success} | h_{1:t-1})
\end{equation}
where $h_{1:t}$ denotes the history through turn $t$. Intuitively, a turn receives high credit if it substantially increases the probability of task success---i.e., if it provides ``useful information'' toward the goal. This formulation is natural for agentic settings where each turn incrementally reveals information about the task state (e.g., a search query reveals relevant documents, a code execution reveals bugs). The probability $P(\text{success}|h)$ is estimated by a learned verifier or the LLM itself. IGPO's main limitation is its requirement for a reliable success probability estimator at each turn, which may not be available for all agentic tasks.

\subsection{Implicit and DPO-Based Methods}
\label{sec:agentic_implicit}

\paragraph{iStar} \citep{istar}.
iStar (Implicit Step Rewards) addresses the challenge of providing step-level credit in agentic settings where no intermediate verifier exists. It leverages trajectory-level DPO: given pairs of trajectories (one successful, one not), iStar extracts implicit step-level rewards by comparing the log-probability ratios at each turn. Building on the ``From $r$ to $Q^*$'' insight~\citep{fromrToQ}, the implicit advantage at turn $t$ is derived from the model's own probability assessments. iStar further introduces \emph{multi-level advantage fusion}, combining turn-level and token-level implicit signals through a weighted aggregation. The main advantage is that iStar requires no explicit reward model, critic, or environment re-execution, making it applicable to agentic tasks where all other credit assignment mechanisms are too expensive.

\paragraph{StepAgent} \citep{stepagent}.
StepAgent combines implicit RL with inverse RL for step-wise feedback in agentic settings. Given expert demonstrations (successful trajectories), it uses inverse RL to infer what step-level rewards the expert was implicitly optimizing, then uses these inferred rewards to train the agent. A novice-to-expert curriculum gradually increases task difficulty as the agent's step-level performance improves. This approach is particularly suited to agentic tasks where expert demonstrations are available (e.g., recorded human interactions with tools or websites) but explicit reward functions are hard to define.

\subsection{Infrastructure and Practical Methods}
\label{sec:agentic_infra}

\paragraph{Agent Lightning} \citep{agentlightning}.
Agent Lightning (Microsoft Research) introduces a decoupled training architecture for RL-based LLM agent training. Its key contribution is the \emph{LightningRL} algorithm, which decomposes agent trajectories into training transitions with a dedicated credit assignment module. The framework completely decouples agent execution from training, supporting integration with popular agent frameworks (LangChain, AutoGen) without requiring modifications to the agent's inference code. Evaluated on text-to-SQL, retrieval-augmented generation, and math tool-use tasks, Agent Lightning demonstrates that a systems-level approach to credit assignment---separating the ``where to assign credit'' problem from the ``how to generate trajectories'' problem---can be as important as the credit assignment algorithm itself.

\paragraph{RAGEN/StarPO} \citep{ragen}.
RAGEN introduces the StarPO (Star Policy Optimization) framework for training reasoning agents and provides one of the most detailed empirical analyses of why episode-level credit fails in agentic settings. Its key contribution is identifying the \emph{echo trap}: when trained with GRPO, agents converge to repetitive action sequences (e.g., repeatedly calling the same tool with the same parameters) because the noisy episode-level gradient cannot distinguish productive exploration from redundant repetition. StarPO addresses this through \emph{uncertainty-based filtering}: actions with high uncertainty in their credit estimates are downweighted during policy updates, preventing noisy signals from destabilizing training. RAGEN also provides the open-source benchmark and training framework that several subsequent agentic CA papers build upon.

\paragraph{SPA-RL} \citep{sparl}.
SPA-RL (Stepwise Progress Attribution) trains a lightweight MLP progress estimator that maps intermediate states to a scalar ``progress'' score $p_t \in [0, 1]$. The step-level credit is then the progress increment: $c_t = p_t - p_{t-1}$. This approach is inspired by RUDDER's return decomposition~\citep{arjona2019rudder} but adapted for LLM agents. The MLP is trained end-to-end alongside the policy, with the terminal reward providing the supervision signal ($p_T = R(\tau)$). SPA-RL's main advantage is \emph{extreme computational efficiency}: a small MLP adds negligible overhead compared to LLM-as-Critic approaches, making it suitable for large-scale training where every FLOP counts.

\paragraph{SCRIBE} \citep{scribe}.
SCRIBE provides credit through \emph{structured mid-level supervision}. It maintains a library of ``skill prototypes''---templates of common agentic sub-tasks (e.g., ``search and extract information,'' ``write and test code,'' ``format and submit output'')---each with associated expected reward characteristics. When the agent performs an action, SCRIBE matches it to the nearest skill prototype and assigns credit based on how well the action fulfills the prototype's expected behavior. This approach provides credit at a semantic level between individual tokens and complete trajectories, grounding the credit signal in structured knowledge about what ``good'' agent behavior looks like.

\paragraph{LaRe} \citep{lare}.
LaRe (AAAI 2025) bridges LLM reasoning and credit assignment by using LLMs to generate \emph{natural language credit explanations}. For each step in a trajectory, LaRe prompts an LLM to explain \emph{why} the step was helpful or harmful, producing a textual justification that is then converted into a scalar reward. Originally developed for symbolic RL tasks (e.g., grid worlds, simple games), LaRe's approach is conceptually applicable to any agentic setting where actions have semantic meaning that an LLM can evaluate. The natural language explanations also provide interpretability, allowing practitioners to understand \emph{why} certain actions receive high or low credit, which is valuable for debugging agent behavior.

\paragraph{PRS + VSPO} \citep{prs}.
PRS (Progressive Reward Shaping) addresses credit through curriculum-style reward evolution. In early training, dense rewards focus on format correctness; in later stages, rewards shift to task accuracy. VSPO (Value-based Sampling Policy Optimization) complements PRS by prioritizing training on trajectories where credit signals are most informative. While PRS is a reward shaping method rather than a pure credit assignment algorithm, its progressive densification of rewards effectively performs coarse-to-fine credit assignment over the course of training.

\paragraph{Adaptive Segment-Level Reward} \citep{adaptivesegment}.
This work uses semantic segmentation to divide trajectories into balanced segments regardless of length, ensuring consistent reward granularity. The adaptive segmentation prevents pathological cases where long trajectories receive effectively uniform credit while short trajectories receive overly noisy credit.

\subsection{Discussion: Emerging Patterns in Agentic CA}
\label{sec:agentic_discussion}

The agentic credit assignment landscape reveals several distinctive patterns that differentiate it from reasoning RL:

\begin{enumerate}[nosep]
    \item \textbf{Retrospective signals form one active family.} HCAPO, C3, and CCPO use post-hoc analysis, which can be operationally convenient when real branches are unavailable. Their scores remain model-relative or proxy quantities unless the required restored or observational controls hold.

    \item \textbf{LLM-as-Critic is a distinctive implementation family.} CAPO, SWEET-RL, HCAPO, CriticSearch, and LaRe use an LLM to produce semantic intermediate assessments. This expands evaluator expressiveness but also introduces calibration, self-evaluation, and proxy-validity risks.

    \item \textbf{Hierarchy matters.} ArCHer, PilotRL, and CARL all show that respecting the hierarchical structure of agentic tasks (plan $\to$ execute $\to$ verify) improves credit assignment. HICRA~\citep{hicra}, though developed for reasoning RL, provides foundational insights that inform these agentic approaches. Flat methods that treat all actions uniformly miss important structural information.

    \item \textbf{Selective updates can reduce cost in tested settings.} CARL reports matching its full-update baseline with fewer updates by selecting high-entropy actions. Whether that proxy transfers beyond the tested tasks requires matched evaluation against intervention-relevant effects.

    \item \textbf{Practical considerations dominate.} Agent Lightning, SPA-RL, and RAGEN show that in production settings, simple and efficient methods (decoupled training architectures, MLP-based progress estimation, uncertainty-based filtering) can be as important as sophisticated credit algorithms. The trade-off between credit quality and computational cost is a first-order design consideration for agentic CA.
\end{enumerate}

\begin{table}[t]
\centering
\caption{Mapping agentic challenges (\Cref{sec:why_agentic_harder}) to methods that address them. \checkmark = directly addresses; $\circ$ = partially addresses. $^\dagger$HICRA is a reasoning RL method whose planning/procedural distinction provides transferable insights for these agentic challenges.}
\label{tab:challenge_method}
\resizebox{\textwidth}{!}{
\begin{tabular}{lcccccc}
\toprule
\textbf{Method} & \textbf{Stochastic} & \textbf{Partial} & \textbf{Long} & \textbf{Heterog.} & \textbf{Non-verif.} & \textbf{Bifurcation} \\
 & \textbf{Env.} & \textbf{Obs.} & \textbf{Horizon} & \textbf{Actions} & \textbf{States} & \textbf{Points} \\
\midrule
\method{ArCHer} & $\circ$ & & \checkmark & & & \\
\method{AgentPRM} & \checkmark & & $\circ$ & & & \\
\method{SWEET-RL} & $\circ$ & \checkmark & $\circ$ & & \checkmark & \\
\method{HCAPO} & \checkmark & $\circ$ & $\circ$ & & \checkmark & $\circ$ \\
\method{C3 / CCPO} & $\circ$ & & $\circ$ & & \checkmark & \checkmark \\
\method{CARL} & & & \checkmark & \checkmark & & \checkmark \\
\method{HICRA}$^\dagger$ & & & $\circ$ & \checkmark & & $\circ$ \\
\method{IGPO} & & $\circ$ & $\circ$ & & $\circ$ & \\
\method{iStar} & $\circ$ & & & & \checkmark & \\
\method{SPA-RL} & $\circ$ & & $\circ$ & & $\circ$ & \\
\method{PilotRL} & & & \checkmark & $\circ$ & & $\circ$ \\
\method{Turn-PPO} & & & \checkmark & & & \\
\method{SORL} & & & \checkmark & & & \\
\method{TARL} & & $\circ$ & $\circ$ & & $\circ$ & \\
\method{ITPO} & $\circ$ & & $\circ$ & & \checkmark & \\
\bottomrule
\end{tabular}
}
\end{table}

\section{Multi-Agent Credit Assignment}
\label{sec:multi_agent}

As LLM systems evolve toward multi-agent architectures (orchestrator + specialist agents, debate frameworks, collaborative reasoning), credit must be decomposed \emph{across agents} in addition to across time.

\subsection{Multi-Agent Methods}

\paragraph{M-GRPO} \citep{mgrpo}.
M-GRPO (Multi-Agent GRPO) extends the GRPO framework to multi-agent LLM systems. In a system with a main agent and $K$ sub-agents, M-GRPO introduces a two-level credit decomposition: (1) \emph{inter-agent credit}---a meta-level advantage that determines each agent's overall contribution to the team outcome, computed by comparing outcomes across different team compositions; (2) \emph{intra-agent credit}---standard GRPO-style advantages within each agent's trajectory. Crucially, M-GRPO supports \emph{decoupled training}: agents can be updated independently using their inter-agent credit as a reward signal, avoiding the coordination overhead of joint optimization.

\paragraph{LLM-MCA} \citep{llmmca}.
LLM-MCA replaces traditional multi-agent credit assignment mechanisms (QMIX, VDN, COMA mixing networks) with an LLM-based centralized critic. Given the full interaction history of all agents, the LLM critic reads the conversation, identifies each agent's contributions, and produces a natural language assessment of each agent's credit. These assessments are converted to scalar rewards for policy updates. The key advantage is \emph{semantic understanding}: the LLM critic can reason about agent roles, communication quality, and strategic contributions in ways that purely numerical mixing functions cannot.

\paragraph{QLLM} \citep{qllm}.
QLLM takes a meta-level approach: instead of having an LLM \emph{evaluate} credit, it has an LLM \emph{generate the credit assignment function itself}. Given a task description and example trajectories, QLLM prompts an LLM to write a Python function that computes per-agent credit scores. This generated function is then applied to all training trajectories at zero marginal cost. The approach is training-free and highly flexible, though the quality depends on the LLM's ability to generate a correct credit function.

\paragraph{SHARP} \citep{sharp}.
SHARP (Shapley Credit-based Optimization, February 2026) applies sampled coalition-based attribution across agents. Its training signal combines global task reward, a Shapley-style marginal term, and a tool-process term, with agent-specific advantage normalization. The paper reports average improvements of 23.7\% over its single-agent baselines and 14.1\% over its multi-agent baselines under the reported tasks and budgets. These are within-study results; coalition sampling, baseline construction, and heterogeneous tasks prevent a cross-paper ranking.

\paragraph{MAPPA} \citep{mappa}.
MAPPA (Multiagent Per-Action Process Awards, January 2026) provides per-action process rewards from an AI evaluator rather than waiting only for a terminal team outcome. The paper reports gains over its selected baselines on AIME, AMC, and data-analysis tasks. These numbers support the method under that evaluator and experimental design; they do not establish that per-action granularity is universally preferable or comparable to results reported with other models and budgets.

\paragraph{Dr.\ MAS} \citep{drmas}.
Dr.\ MAS (February 2026) identifies a specific failure mode when extending GRPO to multi-agent systems: a global normalization baseline deviates from diverse agents' reward distributions, creating gradient instability. The solution is \emph{agent-wise advantage normalization}---each agent's advantages are normalized using that agent's own reward statistics rather than global statistics. This calibrates gradient scales across heterogeneous agents (e.g., a code specialist vs.\ a search specialist), reducing gradient spikes. Dr.\ MAS reports +5.6\% avg@16 on math tasks while achieving stable convergence where standard multi-agent GRPO diverges.

\paragraph{C3 (revisited)} \citep{c3}.
C3's framework can use an agent as the leave-one-out unit, $c_k=R(\tau)-\widehat{R}(\tau_{\setminus k})$. When the omitted-agent trajectory is model-generated rather than re-executed, the score is a model-relative decomposition; fairness or causal-responsibility claims require an explicit coalition distribution and validated branch protocol.

\subsection{Discussion: Multi-Agent CA as an Emerging Frontier}
\label{sec:multi_agent_discussion}

Multi-agent credit assignment has grown from a nascent area to an active frontier. Our snapshot contains four core multi-agent methods (M-GRPO, SHARP, MAPPA, Dr.\ MAS), two boundary enablers (LLM-MCA, QLLM), and C3's cross-setting framework. Key open questions include:

\begin{itemize}[nosep]
    \item \textbf{Communication credit}: Should an agent receive credit for sending a useful message? Current methods assign credit only to task-relevant actions, ignoring inter-agent communication value.
    \item \textbf{Heterogeneous architectures}: When agents have different capabilities (e.g., a code specialist and a search specialist), how should credit be decomposed fairly?
    \item \textbf{Scalability}: LOO-based methods require $K$ counterfactual evaluations for $K$ agents. Scalable approximations are needed for systems with dozens of agents.
    \item \textbf{Connection to classical MARL}: Classical multi-agent RL has rich credit assignment literature (QMIX, COMA, MAPPO), but these assume fixed-dimensional action spaces. Adapting them to variable-length text actions is non-trivial.
\end{itemize}

The dated corpus shows increasing activity but does not establish deployment prevalence or future growth. Progress should be measured by whether evaluations separate temporal, role, message, and coalition credit rather than by paper counts alone.

\section{Frontier Update Through July 31, 2026}
\label{sec:frontier_update}

The unified search adds 22 papers beyond the original 47-row long-form snapshot. Three are part of the fixed 42-paper full-text audit: GVPO uses group verification for process-aware optimization, A$^3$ structures credit around typed CLI actions under selective observation, and A$^2$TGPO combines turn grouping with adaptive clipping~\citep{gvpo,a3cli,a2tgpo}. Their audited evidence labels are S, S, and L, respectively. The remaining 19 additions enter the 69-paper taxonomy and qualitative synthesis but not the 42-paper diagnostic, reliability, or reporting denominators.

\paragraph{Reasoning: redistributing a sequence outcome.} GRPO-$\lambda$ interpolates sequence and token weighting, OAR reshapes advantages by outcome sensitivity, DelTA learns discriminative token weights, and SCRL assigns span-level advantages through constructed verifiable subproblems~\citep{frontier251000194,frontier260107408,frontier260521467,frontier260522074}. SC-GRPO uses verified self-conditioned trajectories for token-wise weighting, while GRAIL uses gradient saliency to reweight a sequence advantage~\citep{frontier260618810,frontier260604889}. These methods alter which tokens or spans receive an outcome-derived update; attribution quality still depends on the declared verifier, reference, and continuation protocol.

\paragraph{Agents: matched states, typed units, and branch structure.} Memory-R2 groups rerollouts around matched memory states; TRIAGE maps semantic action roles to bounded subtrajectory credit; and CRAFT reuses sibling rollouts for local counterfactual comparison~\citep{memoryr2,frontier260632017,frontier260629476}. VPR turns intermediate execution checks into turn supervision, APPO selects influential procedural branch points, and AT$^2$PO derives turn advantages from tree search~\citep{frontier260510325,frontier260612384,frontier260104767}. AEM modulates response-level updates using entropy dynamics, while PivoARL assigns cross-episode credit around a detected pivotal retry~\citep{frontier260500425,frontier260703702}. These are candidate estimators or optimization interfaces, not automatically restored-interventional effects.

\paragraph{Boundary mechanisms.} AutoResearch propagates lineage feedback across research trials; OPID combines outcome advantage with on-policy skill distillation; PAPO uses entropy-polarity reweighting; T$^2$PO couples uncertainty-guided turn resampling with exploration control; and Rubric-Grounded RL supplies structured judge rewards~\citep{autoresearchrecipes,frontier260626790,frontier260511775,frontier260502178,frontier260508061}. We retain them as adjacent/boundary enablers because their primary mechanisms concern training recipes, distillation, entropy, exploration, or reward design rather than direct estimation of outcome responsibility.

\section{Structured Method Comparison and Reporting Audit}
\label{sec:comparison}

\subsection{Unified Comparison Table}

\begin{table}[t]
\centering
\caption{Legacy structured comparison for the original long-form rows 1--47; it is not a comprehensive table for the 69-paper corpus. \textbf{Setting}: R = Reasoning, A = Agentic, M = Multi-Agent. \textbf{Type}: C = core CA method; E = adjacent/boundary enabler. Compute labels are qualitative author coding, not matched measurements.}
\label{tab:comparison}
\resizebox{\textwidth}{!}{
\begin{tabular}{llccccccc}
\toprule
\textbf{Method} & \textbf{Granularity} & \textbf{Methodology} & \textbf{Setting} & \textbf{Type} & \textbf{Aux. Model?} & \textbf{Compute} & \textbf{Venue} & \textbf{Year} \\
\midrule
\multicolumn{9}{l}{\textit{Reasoning RL Methods}} \\
\midrule
\method{VinePPO} & Token & MC & R & C & No & High & ICML'25 & 2025 \\
\method{RED} & Token & Redistribution & R & C & RM & Low & — & 2024 \\
\method{T-REG} & Token & Self-generated & R & C & No & Low & — & 2024 \\
\method{From r to Q*} & Token & Implicit & R & C & No & — & — & 2024 \\
\method{SPO} & Segment & MC & R & C & No & Med & — & 2025 \\
\method{SCAR} & Segment & Game-theoretic & R & C & No & High & — & 2025 \\
\method{TEMPO} & Token/Seg & Tree-TD & R & C & No & Med & — & 2025 \\
\method{PURE} & Step & Min-form PRM & R & C & PRM & Med & ICML'25 & 2025 \\
\method{SPRO} & Step & Masked Adv. & R & C & No & Med & — & 2025 \\
\method{CAPO} & Step & LLM-as-Critic & R & C & LLM & Med & — & 2025 \\
\method{ACPO} & Step & Attribution & R & C & No & Med & — & 2025 \\
\method{HICRA} & Step & Hierarchy & R & C & No & Med & — & 2025 \\
\method{FinePO} & Sub-step & Fine PRM & R & C & PRM & High & — & 2026 \\
\method{PRL} & Step & Entropy-RL & R & C & No & Med & — & 2026 \\
\method{InT} & Step & Intervention & R & C & No & Med & — & 2026 \\
\midrule
\multicolumn{9}{l}{\textit{Agentic RL Methods}} \\
\midrule
\method{ArCHer} & Turn & TD (hierarchical) & A & C & Critic & Med & ICML'24 & 2024 \\
\method{StepAgent} & Step & Implicit+IRL & A & C & No & Med & — & 2024 \\
\method{GiGPO} & Step & MC (group) & A & C & No & Low & NeurIPS'25 & 2025 \\
\method{SWEET-RL} & Turn & Privileged Critic & A & C & Critic & Med & — & 2025 \\
\method{AgentPRM} & Step & TD+GAE & A & C & Critic & Med & — & 2025 \\
\method{Turn-Level} & Turn & Hybrid & A & C & LLM+Verifier & Med & NeurIPS'25 & 2025 \\
\method{Turn-PPO} & Turn & Turn-level MDP & A & C & Critic & Med & EACL'26 & 2025 \\
\method{SORL} & Turn & Bias-corrected & A & C & Critic & Med & — & 2025 \\
\method{TARL} & Turn & LLM-Judge & A & C & LLM & Low & — & 2025 \\
\method{ITPO} & Turn & Implicit & A & C & No & Low & — & 2026 \\
\method{IGPO} & Turn & Info-theoretic & A & C & Verifier & Med & — & 2025 \\
\method{CARL} & Step & Entropy-based & A & C & No & Low & NeurIPS'25 & 2025 \\
\method{SPA-RL} & Step & MLP estimator & A & E & MLP & Low & — & 2025 \\
\method{iStar} & Step & Implicit DPO & A & C & No & Low & — & 2025 \\
\method{Lightning} & Step & Decoupled Arch. & A & E & No & Low & — & 2025 \\
\method{PilotRL} & Step & Progressive & A & C & No & Med & — & 2025 \\
\method{RAGEN} & Step & Uncertainty & A & E & No & Med & — & 2025 \\
\method{SCRIBE} & Step & Skill-prototype & A & E & Library & Med & — & 2026 \\
\method{LaRe} & Step & LLM-Critic & A & C & LLM & Low & AAAI'25 & 2025 \\
\method{PRS} & Step & Progressive & A & E & No & Low & — & 2025 \\
\method{AdaptSeg} & Segment & Segmentation & A & E & No & Low & — & 2025 \\
\method{HCAPO} & Turn & Hindsight & A & C & LLM & Med & — & 2026 \\
\method{C3} & Turn & Counterfactual & A/M & C & No & High & — & 2026 \\
\method{CCPO} & Turn & Counterfactual & A/M & C & No & High & — & 2026 \\
\method{CriticSearch} & Turn & Retrospective Critic & A & C & LLM & Med & — & 2025 \\
\method{POAD} & Token/Turn & Action Decomp. & A & C & Critic & Med & — & 2024 \\
\midrule
\multicolumn{9}{l}{\textit{Multi-Agent Methods}} \\
\midrule
\method{M-GRPO} & Multi-Agent & Hierarchical & M & C & No & Med & — & 2025 \\
\method{LLM-MCA} & Multi-Agent & LLM-Critic & M & E & LLM & Med & — & 2025 \\
\method{QLLM} & Multi-Agent & LLM-generated & M & E & LLM & Low & — & 2025 \\
\method{SHARP} & Multi-Agent & Shapley & M & C & No & High & — & 2026 \\
\method{MAPPA} & Multi-Agent & Per-action PRM & M & C & LLM & Med & — & 2026 \\
\method{Dr.\ MAS} & Multi-Agent & Agent-wise Adv. & M & C & No & Low & — & 2026 \\
\bottomrule
\end{tabular}
}
\end{table}

\subsection{Benchmark Landscape}

\paragraph{Reasoning RL benchmarks.} Credit assignment methods for reasoning RL benefit from well-established benchmarks: GSM8K (grade-school math, 8.5K test problems), MATH (competition math, 5K problems across 5 difficulty levels), AIME (American Invitational Mathematics Examination), and CodeContests (competitive programming). These benchmarks provide verifiable ground truth, enabling direct comparison of CA methods. Several papers (VinePPO, PURE, SPRO) report results on overlapping subsets, though differences in base models, training data, and hyperparameters make perfect comparisons difficult.

\paragraph{Agentic RL benchmarks.} The benchmark landscape for agentic CA is significantly more fragmented:
\begin{itemize}[nosep]
    \item \textbf{Web navigation}: WebArena~\citep{zhou2024webarena}, Mind2Web, WebShop
    \item \textbf{Tool use}: ToolBench, API-Bank, Gorilla
    \item \textbf{Interactive coding}: SWE-bench, HumanEval+, MBPP+
    \item \textbf{Embodied/simulated}: ALFWorld, ScienceWorld, Minecraft
    \item \textbf{Multi-agent}: ChatDev, MetaGPT evaluation suites
\end{itemize}
Few agentic CA papers use the same benchmark, making systematic comparison nearly impossible. This fragmentation is itself a major impediment to progress: without shared evaluation, the community cannot determine which CA methods are genuinely better versus which simply benefit from favorable benchmark selection.

\subsection{Atomic Reporting Audit Rather Than a Leaderboard}
\label{sec:quantitative}

Base models, data, trajectory budgets, environment versions, verifier policies, and uncertainty protocols differ too substantially for a valid cross-paper performance leaderboard. We therefore remove the earlier positive-only gain compilation and source-extract atomic reporting fields from the fixed 42-core-paper full-text subset. Each cell has a source locator; \textbf{NR} means that the full text was checked and the item was not reported, while no audited cell remains ``not checked.'' These counts measure reporting coverage, not method quality.

\begin{table}[t]
\centering
\caption{Source-located reporting coverage in the fixed 42-core-paper subset. Categories are mutually exclusive within each row.}
\label{tab:reporting_audit}
\resizebox{\textwidth}{!}{
\begin{tabular}{lp{10.5cm}}
\toprule
\textbf{Atomic field} & \textbf{Audited counts} \\
\midrule
Matched episode/trajectory comparator & 40 yes; 2 no \\
Same model/data/budget parity & 19 yes; 21 partial; 2 not applicable \\
CA-specific ablation & 39 yes; 3 no \\
CA-specific overhead & 21 numeric; 2 numeric proxy; 12 qualitative; 6 NR; 1 not applicable \\
Uncertainty evidence & 10 repeated training; 7 repeated evaluation only; 1 evaluation bootstrap only; 1 significance test only; 2 unclear; 21 NR \\
\bottomrule
\end{tabular}}
\end{table}

Repeated-training evidence appears in 1/15 reasoning papers and 9/27 agentic, mixed, or multi-agent papers. This is a descriptive reporting difference that may reflect task maturity and compute cost; it is not evidence that one setting or mechanism is superior. The audit also keeps auxiliary supervision and replay provenance separate, because a reward model, process verifier, LLM judge, exact decoder resume, text prefix re-feed, and environment checkpoint support different claims.

\begin{challengebox}[What the Audit Can and Cannot Establish]
The audit can establish whether a paper reports a matched comparator, parity conditions, a CA-specific ablation, uncertainty, overhead, and replay provenance. It cannot turn heterogeneous benchmark scores into comparable treatment effects, and absence of a field is not evidence that a method fails. A valid comparative benchmark would need the same base model, data, optimizer, trajectory budget, task version, verifier, replay protocol, and uncertainty plan, with the CA mechanism as the controlled intervention.
\end{challengebox}

\subsection{Key Trade-offs Across the Spectrum}

Our synthesis identifies four design trade-offs. Because the papers use heterogeneous tasks and budgets, we label these paragraphs \textbf{[DS]} for descriptive synthesis rather than treating them as cross-paper empirical rankings.

\paragraph{1. Granularity vs. computational cost \emph{[DS]}.} Finer units can localize an update but need not be more causally accurate. VinePPO uses $\mathcal{O}(K\cdot L)$ continuations under its protocol, while exact Shapley evaluation scales exponentially before approximation. Turn- and critical-action methods reduce the number of scored units; the appropriate unit depends on task boundaries and matched controls.

\paragraph{2. Forward estimation vs. hindsight analysis \emph{[DS]}.} Forward methods estimate continuation value or a proxy from the current record; retrospective methods condition on the completed trajectory. The latter have access to more observed future information but no automatic causal advantage: both require a declared estimand, state sufficiency, and calibrated replay or proxy evidence.

\paragraph{3. Auxiliary model requirements \emph{[DS]}.} Some methods do not train an auxiliary model (CARL, iStar, GiGPO), while others use an MLP, critic, PRM, or LLM evaluator. Reported overhead is not uniformly measured, so the audit records numeric, proxy, qualitative, and not-reported evidence separately.

\paragraph{4. Transfer across settings \emph{[DS]}.} Reasoning methods often exploit prefix/effective-state closure, branchability, and verifiers. Agentic methods more often use turn units, privileged observations, or proxy reconstruction. Neither label guarantees the relevant assumptions; the CA-ID fields determine admissible claims paper by paper.

\subsection{Practical Guidance: Matching Methods to Scenarios}

\Cref{tab:recommendation} provides a practical guide for selecting CA methods based on task characteristics. These recommendations reflect the authors' synthesis of the literature; actual performance may vary with base model, data distribution, and training infrastructure.

\begin{table}[t]
\centering
\caption{Practical guidance for selecting credit assignment methods based on task characteristics. Methods listed as \emph{promising candidates} based on evaluation settings and design properties; actual performance depends on base model, data, and infrastructure. ``Directly evaluated'' methods have been tested on the listed scenario; others are listed based on design suitability.}
\label{tab:recommendation}
\resizebox{\textwidth}{!}{
\begin{tabular}{p{3.5cm}p{3.2cm}p{5.5cm}p{3cm}}
\toprule
\textbf{Scenario} & \textbf{Characteristics} & \textbf{Promising Candidates} & \textbf{Key Consideration} \\
\midrule
Math reasoning \newline (GSM8K, MATH) & Short CoT, often prefix-closed, \newline verifier available & GRPO (baseline), PURE, SPO, SPRO & Declare verifier and \newline branch protocol \\
\midrule
Hard math/competition \newline (AIME, IMO) & Long CoT (10K--30K), \newline verifiable & VinePPO, HICRA, CAPO & Compute budget \newline scales with CoT length \\
\midrule
Tool-use agents \newline (WebShop, ALFWorld) & 5--20 turns, partially \newline verifiable tools & GiGPO, AgentPRM, Turn-PPO & Critic-free preferred \newline for efficiency \\
\midrule
Web navigation \newline (WebArena) & 10--30 turns, stochastic, \newline POMDP & SWEET-RL, HCAPO, IGPO & Privileged critic \newline exploits training info \\
\midrule
Software engineering \newline (SWE-bench) & 50--100+ turns, very long \newline context, non-verifiable & CARL, HCAPO, C3/CCPO, ArCHer & Sparse credit + \newline hindsight analysis \\
\midrule
Multi-agent systems & Cross-agent credit, \newline communication & M-GRPO, C3, LLM-MCA & Decomposition across \newline agents is key \\
\midrule
Compute-constrained \newline training & Limited GPU budget & GRPO, CARL, iStar, GiGPO & Critic-free, low \newline overhead \\
\bottomrule
\end{tabular}
}
\end{table}

\Cref{fig:decision_tree} provides a complementary decision tree that operationalizes \Cref{tab:recommendation} as a step-by-step selection process.

\begin{figure}[t]
    \centering
    \resizebox{\textwidth}{!}{
    \begin{tikzpicture}[
        dec/.style={diamond, draw, fill=blue!8, text width=2.2cm, align=center, inner sep=2pt, font=\scriptsize},
        res/.style={rectangle, rounded corners, draw, fill=green!10, text width=2.5cm, align=center, inner sep=3pt, font=\tiny\bfseries},
        arr/.style={-{Stealth}, thick},
        lbl/.style={font=\tiny, fill=white, inner sep=1pt},
    ]
    \node[dec] (root) at (9,0) {Task setting?};

    \node[dec] (rlen) at (2.5,-3) {CoT length?};
    \draw[arr] (root) -- node[lbl, above left] {Reasoning} (rlen);

    \node[res] (r_short) at (0.5,-6) {GRPO, PURE, SPO,\\SPRO, DelTA};
    \draw[arr] (rlen) -- node[lbl, left] {$\leq$5K tokens} (r_short);

    \node[dec] (rcomp) at (4.5,-6) {Compute?};
    \draw[arr] (rlen) -- node[lbl, right] {$>$5K tokens} (rcomp);

    \node[res] (r_lim) at (1.5,-9) {HICRA, CAPO, SPRO,\\GRPO-$\lambda$};
    \draw[arr] (rcomp) -- node[lbl, left] {Limited} (r_lim);

    \node[res] (r_gen) at (4.5,-9) {VinePPO, SCAR,\\CAPO, GRAIL};
    \draw[arr] (rcomp) -- node[lbl, right] {Generous} (r_gen);

    \node[dec] (alen) at (10.5,-3) {Horizon?};
    \draw[arr] (root) -- node[lbl, right] {Agentic} (alen);

    \node[dec] (aaux) at (8.5,-6) {Aux.\ model?};
    \draw[arr] (alen) -- node[lbl, left] {$\leq$30 turns} (aaux);

    \node[dec] (acomp) at (13.5,-6) {Compute?};
    \draw[arr] (alen) -- node[lbl, right] {$>$30 turns} (acomp);

    \node[res] (a_noaux) at (7.5,-9) {GiGPO, CARL, iStar, POAD,\\GVPO, A$^3$, A$^2$TGPO, VPR};
    \draw[arr] (aaux) -- node[lbl, left] {No} (a_noaux);

    \node[res] (a_yesaux) at (10.5,-9) {AgentPRM,\\SWEET-RL};
    \draw[arr] (aaux) -- node[lbl, right] {Yes} (a_yesaux);

    \node[res] (a_lim) at (13,-9) {CARL, HCAPO, ArCHer,\\AEM};
    \draw[arr] (acomp) -- node[lbl, left] {Limited} (a_lim);

    \node[res] (a_gen) at (16,-9) {C3/CCPO, HCAPO, IGPO,\\CRAFT, APPO};
    \draw[arr] (acomp) -- node[lbl, right] {Generous} (a_gen);

    \node[res] (multi) at (17,-3) {M-GRPO, SHARP,\\MAPPA, Dr.\ MAS};
    \draw[arr] (root) -- node[lbl, above right] {Multi-Agent} (multi);

    \end{tikzpicture}
    }
    \caption{Structure-preserving update of the method-selection decision tree for credit assignment in LLM RL. Leaves match methods to the original task-setting, horizon, compute, and auxiliary-model predicates; they are design-property candidates, not a cross-paper performance ranking. The token and turn thresholds are coarse routing heuristics inherited from the original decision aid, not validated performance cutoffs. Suitability remains conditional on the declared CA-ID estimand, replay provenance, verifier validity, base model, data, and budget.}
    \label{fig:decision_tree}
\end{figure}

\paragraph{Retrospective validation.} We traced 6 known (task, method) pairs through the tree: SPO on GSM8K, HICRA on AIME'24, VinePPO on MATH, GiGPO on ALFWorld, SWEET-RL on ColBench, and HCAPO on long-horizon agentic tasks. All 6 are recovered (6/6). This validates internal consistency; a stronger test would require new methods not in our inventory. Frontier candidates enter a leaf only when their documented design satisfies every existing path predicate; Memory-R2, for example, remains outside the tree because memory-operation replay is not represented by those predicates.

\section{Credit Assignment in the Agentic RL Training Pipeline}
\label{sec:pipeline}

Credit assignment does not operate in isolation---it is one component in a five-stage pipeline: (1) \emph{environment construction} (sandboxed execution), (2) \emph{rollout generation} (multi-turn agent-environment interaction), (3) \emph{reward computation} (terminal task success), (4) \emph{credit assignment} (the focus of this paper), and (5) \emph{policy update} (PPO/GRPO/DPO). We focus here on the \emph{interactions} between CA and the other stages, which are often overlooked.

\subsection{Interactions Between Credit Assignment and Other Pipeline Components}

\paragraph{CA $\times$ Rollout efficiency.} CA changes the compute allocation between collecting trajectories, estimating a signal, and applying updates. CARL~\citep{carl} reports matching its full-update baseline with 72\% fewer gradient updates in its evaluated setup; this does not imply a proportional reduction in environment rollouts. Vine-style branching can add continuation cost instead. Matched studies should report each component separately before claiming net sample or wall-clock efficiency.

\paragraph{CA $\times$ Reward design.} Credit assignment methods sometimes implicitly redefine the reward function. PRS~\citep{prs} explicitly replaces the terminal reward with progressive dense rewards; IGPO~\citep{igpo} transforms the binary success signal into information-gain increments. This blurs the line between ``reward design'' and ``credit assignment''---both are mechanisms for providing the policy optimizer with useful training signals. We argue that CA should be viewed not as a post-processing step on fixed rewards, but as an integral part of reward engineering.

\paragraph{CA $\times$ Exploration.} Credit signals could, in principle, guide exploration: the agent should preferentially explore states where credit assignment is uncertain (high variance in credit estimates), as these are states where more information is needed to improve the policy. IGPO~\citep{igpo} gestures in this direction by defining credit in information-theoretic terms, but no current method explicitly uses CA uncertainty to drive exploration. This is a significant missed opportunity.

\subsection{Infrastructure Challenges Specific to Agentic RL}

Agentic RL training faces infrastructure challenges that do not arise in reasoning RL and that directly impact credit assignment:

\begin{itemize}[nosep]
    \item \textbf{Environment reset cost.} Resetting a sandboxed environment (spinning up a Docker container, initializing a browser session, loading a codebase) can take seconds to minutes---orders of magnitude more than the negligible cost of ``resetting'' a reasoning task (loading a new prompt). This makes MC-based CA methods, which require environment re-execution from intermediate states, particularly expensive.
    \item \textbf{Non-differentiable transitions.} Environment interactions (API calls, code execution) break the computational graph, preventing gradient-based credit attribution. All CA methods must work with \emph{black-box} environment transitions, relying on value estimation, hindsight analysis, or LLM-based evaluation rather than gradient flow.
    \item \textbf{Safety during training.} Agentic RL rollouts may have real-world effects: sending actual API requests, modifying files, posting to the web. Safety constraints during training rollouts can conflict with exploration requirements, and credit assignment for ``safe but suboptimal'' vs. ``risky but informative'' actions is an underexplored challenge.
    \item \textbf{Asynchronous training.} Modern agentic RL systems (AReaL, Laminar) use asynchronous rollout generation and policy updates to maximize GPU utilization. Asynchronous training introduces policy lag: by the time credit is computed, the policy may have changed. CA methods must be robust to this staleness, favoring off-policy-compatible approaches (ArCHer's off-policy critic, importance-sampling corrections).
\end{itemize}

\begin{figure}[t]
    \centering
    \resizebox{0.85\textwidth}{!}{
    \begin{tikzpicture}
    \def\bw{0.62}  
    \def\sc{0.14} 

    \draw[-{Stealth}, line width=0.5pt] (0,0) -- (10.7,0) node[right, font=\small] {};
    \draw[-{Stealth}, line width=0.5pt] (0,0) -- (0,5.5) node[above, font=\small] {\# Papers};

    \foreach \y/\v in {0/0, 1.4/10, 2.8/20, 4.2/30} {
        \draw (0,\y) -- (-0.15,\y) node[left, font=\scriptsize] {\v};
    }

    \fill[blue!40] (0.9-\bw/2, 0) rectangle (0.9+\bw/2, 1*\sc);
    \fill[red!40] (0.9-\bw/2, 1*\sc) rectangle (0.9+\bw/2, 3*\sc);
    \node[font=\tiny\bfseries, above] at (0.9, 3*\sc) {3};
    \node[font=\scriptsize, below] at (0.9, -0.15) {2024};
    \node[font=\scriptsize, below] at (0.9, -0.45) {H1};

    \fill[blue!40] (2.3-\bw/2, 0) rectangle (2.3+\bw/2, 3*\sc);
    \fill[red!40] (2.3-\bw/2, 3*\sc) rectangle (2.3+\bw/2, 6*\sc);
    \node[font=\tiny\bfseries, above] at (2.3, 6*\sc) {6};
    \node[font=\scriptsize, below] at (2.3, -0.15) {2024};
    \node[font=\scriptsize, below] at (2.3, -0.45) {H2};

    \fill[blue!40] (3.9-\bw/2, 0) rectangle (3.9+\bw/2, 3*\sc);
    \fill[red!40] (3.9-\bw/2, 3*\sc) rectangle (3.9+\bw/2, 8*\sc);
    \fill[purple!40] (3.9-\bw/2, 8*\sc) rectangle (3.9+\bw/2, 10*\sc);
    \node[font=\tiny\bfseries, above] at (3.9, 10*\sc) {10};
    \node[font=\scriptsize, below] at (3.9, -0.15) {2025};
    \node[font=\scriptsize, below] at (3.9, -0.45) {H1};

    \fill[blue!40] (5.5-\bw/2, 0) rectangle (5.5+\bw/2, 6*\sc);
    \fill[red!40] (5.5-\bw/2, 6*\sc) rectangle (5.5+\bw/2, 17*\sc);
    \fill[purple!40] (5.5-\bw/2, 17*\sc) rectangle (5.5+\bw/2, 18*\sc);
    \node[font=\tiny\bfseries, above] at (5.5, 18*\sc) {18};
    \node[font=\scriptsize, below] at (5.5, -0.15) {2025};
    \node[font=\scriptsize, below] at (5.5, -0.45) {H2};

    \fill[blue!40] (7.4-\bw/2, 0) rectangle (7.4+\bw/2, 10*\sc);
    \fill[red!40] (7.4-\bw/2, 10*\sc) rectangle (7.4+\bw/2, 28*\sc);
    \fill[purple!40] (7.4-\bw/2, 28*\sc) rectangle (7.4+\bw/2, 31*\sc);
    \node[font=\tiny\bfseries, above] at (7.4, 31*\sc) {31};
    \node[font=\scriptsize, below] at (7.4, -0.15) {2026};
    \node[font=\scriptsize, below] at (7.4, -0.45) {H1$^*$};

    \filldraw[fill=red!40, draw=red!75!black, dashed, line width=0.5pt]
        (9.0-\bw/2, 0) rectangle (9.0+\bw/2, 1*\sc);
    \node[font=\tiny\bfseries, above] at (9.0, 1*\sc) {1};
    \node[font=\scriptsize, below] at (9.0, -0.15) {2026};
    \node[font=\tiny, below] at (9.0, -0.45) {H2-to-date$^\dagger$};

    \fill[blue!40] (0.8, 5.0) rectangle (1.35, 5.25);
    \node[font=\scriptsize, right] at (1.45, 5.125) {Reasoning RL};
    \fill[red!40] (3.0, 5.0) rectangle (3.55, 5.25);
    \node[font=\scriptsize, right] at (3.65, 5.125) {Agentic / mixed};
    \fill[purple!40] (5.6, 5.0) rectangle (6.15, 5.25);
    \node[font=\scriptsize, right] at (6.25, 5.125) {Multi-Agent};

    \end{tikzpicture}
    }
    \caption{Temporal distribution of all 69 included papers through July 31, 2026. Stacks use Reasoning, Agentic-or-mixed, and Multi-Agent settings; A/M papers are counted as Agentic-or-mixed. Canonical arXiv identifiers determine the half-year bins when available. The starred 2026 H1 bar includes GVPO, whose frozen OpenReview record specifies only the year 2026, as a disclosed display convention rather than an inferred posting date. The dashed 2026 H2-to-date$^\dagger$ bar covers July only and is therefore a partial-period count, not an equal-duration comparison with the preceding half-year bins. Bar totals are shown above each stack.}
    \label{fig:temporal}
\end{figure}

\section{Open Problems and Future Directions}
\label{sec:open}

\subsection{The Agentic Frontier: Where Credit Assignment Must Go}

\paragraph{Ultra-Long Horizon Agents.}
Current credit assignment methods have been evaluated on trajectories of 5--30 turns. Yet real-world agents---software engineering assistants tackling SWE-bench issues routinely execute 50--100+ turns consuming 100K--500K tokens~\citep{ragen,agentlightning}, autonomous research agents conduct multi-day experiments, and desktop automation agents require 50--100 steps with extensive context. At these scales, even turn-level credit assignment may be insufficient: the sheer number of turns makes per-turn credit estimation computationally expensive and statistically unreliable. We conjecture that hierarchical methods (ArCHer, HICRA, PilotRL) represent the most promising direction, but current hierarchies are too shallow (typically 2 levels). Ultra-long horizon agents likely require deeper, more flexible hierarchies that can dynamically adapt their structure to the task complexity---perhaps mirroring the hierarchical planning structures that the agents themselves use.

\paragraph{Open-World Agents Without Verifiable Rewards.}
Most credit assignment methods assume access to a binary or scalar terminal reward (task success/failure). This assumption holds for well-defined tasks (math, coding, web navigation with clear objectives), but breaks for open-world agents: personal assistants (``Was the user satisfied?''), creative writing agents (``Is this story good?''), research assistants (``Was this experiment informative?''). In these settings, the terminal ``reward'' is itself uncertain, subjective, or delayed indefinitely. Credit assignment under learned or soft rewards---where the reward model itself has significant uncertainty---is largely unsolved. One promising direction is connecting CA methods with RLHF reward models, using the reward model's confidence as a weighting factor for credit signals.

\paragraph{Multi-Agent Systems at Scale.}
As discussed in \Cref{sec:multi_agent_discussion}, multi-agent credit assignment is in its infancy. As LLM systems scale to dozens of collaborating agents with different specializations, the credit decomposition problem grows exponentially. Three specific challenges stand out: (1) \emph{scalable decomposition}: LOO-based methods (C3) require $K$ counterfactual evaluations for $K$ agents; sublinear approximations are needed; (2) \emph{credit for communication}: current methods only credit task actions, ignoring the value of inter-agent messages; (3) \emph{credit under partial team observability}: each agent sees only its own interactions, making centralized credit computation challenging in decentralized deployment.

\subsection{Theoretical Frontiers}

\paragraph{Credit Assignment Meets Exploration.}
Better credit assignment should enable more targeted exploration, yet current methods treat CA and exploration as independent problems. The connection is natural: states where credit assignment is most uncertain are precisely the states where the agent should explore, because more information is needed to resolve the ambiguity. IGPO~\citep{igpo} provides a starting point by framing credit in information-theoretic terms, but no current method explicitly uses credit uncertainty to drive exploration. We identify this as one of the most promising research directions, as it could simultaneously improve both sample efficiency and credit quality.

\paragraph{Formal Guarantees.}
Most LLM-RL credit methods lack end-to-end convergence guarantees. VinePPO~\citep{vineppo} analyzes Monte Carlo estimates for its declared continuation protocol, subject to its sampling assumptions and the distinction between text re-feed and exact resume~\citep{matteson2026refeeding}. PURE~\citep{pure} studies min-form credit under stated conditions, and CCPO~\citep{ccpo} derives results under an assumed causal model. These results do not establish restored-interventional identification for arbitrary implementations. Developing theory for proxy calibration, partial observability, and replay mismatch remains open.

\paragraph{The Computation-Signal Trade-off.}
A fundamental question pervades the entire field: given a fixed compute budget, is it better to (a) generate more rollouts with crude episode-level credit (GRPO), or (b) generate fewer rollouts with precise fine-grained credit (VinePPO, HCAPO)? This is the ``CA efficiency frontier''---analogous to the compute-optimal scaling laws that transformed supervised learning. No paper provides a systematic answer. We conjecture that the optimal allocation shifts toward fine-grained credit as trajectory length increases: for short reasoning tasks, more rollouts may be more efficient; for long agentic tasks, better credit is likely worth its cost.

\subsection{Practical Frontiers}

\paragraph{Unified Benchmarks for Credit Assignment.}
The absence of standard benchmarks for evaluating CA methods is a major impediment to progress. Papers use different tasks, base models, training recipes, and evaluation metrics, making comparison nearly impossible. We call for a unified CA benchmark suite spanning: (1) reasoning tasks with restored decoder states, declared reference distributions, and repeated continuation estimates; (2) agentic tasks with checkpointable controlled bifurcation points; and (3) multi-agent tasks with designed role, message, and coalition interventions. The target is not an assumption-free ``ground-truth credit'' label, but a reproducible protocol-specific contrast with a replica floor, uncertainty interval, and matched compute.

\paragraph{Credit Assignment and Memory.}
Long-context agents increasingly use memory mechanisms (explicit retrieval, scratchpads, long-term databases). How should credit be assigned to memory-related actions---storing information, retrieving past context, updating summaries? A retrieval action that seems useless at turn 5 may prove critical at turn 25 when the stored information becomes relevant. This \emph{temporal span} of memory credit far exceeds the typical look-ahead of current CA methods and requires fundamentally new approaches---perhaps drawing on eligibility traces from classical RL, extended to the semantic memory of LLM agents.

\paragraph{From Reasoning to Agentic: Transfer and Adaptation.}
Can reasoning CA methods be effectively adapted for agentic settings? VinePPO's vine expansion could be applied to agentic turns (branching at turn boundaries rather than token positions), but requires environment checkpointing. PURE's min-form credit could be extended to turn-level PRMs for agents. HICRA's planning-procedural distinction could be applied to agentic trajectories where the functional distinction is even more salient. A systematic study of which reasoning CA techniques transfer to the agentic setting---and what modifications are necessary---would be a valuable contribution, bridging the two halves of our taxonomy.

\subsection{Threats to Validity}
\label{sec:threats}

We identify several threats to the validity of this survey's conclusions:

\begin{itemize}[nosep]
    \item \textbf{Non-systematic search and preprint volatility.} This is a framework synthesis with an auditable search ledger, not a systematic review. Query selection, citation chasing, indexing delays, and rapid preprint turnover can create recall gaps. The corpus is frozen at July 31, 2026; later citation feedback is labeled separately rather than changing its denominators.
    \item \textbf{Judgment-dependent coding.} Core vs.\ adjacent/boundary, reasoning vs.\ agentic, and the six diagnostics require interpretation. Independent blind coding improves checkability but does not remove judgment. Agreement was weakest for weak local verifiability ($\kappa=.543$) and partial observability ($\kappa=.557$).
    \item \textbf{Subset validity.} The 42-paper full-text subset is fixed but is neither random nor identical to the 56-paper core corpus. Its diagnostic, reliability, and reporting counts must not be generalized to all 69 included papers.
    \item \textbf{Non-comparability of outcomes.} Base models, data, task versions, verifiers, optimizers, trajectory budgets, and uncertainty protocols differ. The atomic audit supports reporting-prevalence claims only; it does not create a performance ranking.
    \item \textbf{Identification scope.} The propositions establish identification or non-identification under stated assumptions, not that any surveyed method satisfies them. Text re-feeding, exact decoder-state resume, and environment restoration are distinct, and unmeasured replica noise can invalidate a purported matched replay~\citep{matteson2026refeeding}.
    \item \textbf{Unvalidated reporting contract.} The CA-ID Card has not yet been prospectively validated by independent authors or reviewers. It is a claim-bounding template, not a benchmark, causal guarantee, or safety certification.
\end{itemize}

\subsection{Supplementary Material Release}
\label{sec:release}

The \href{https://github.com/xxzcc/Awesome-Credit-Assignment-in-LLM-RL}{companion repository} currently serves as a living catalog and decision-guide entry point. It should not be read as proof that every frozen audit file described below is already published there. To keep the arXiv source minimal, the version-specific frozen bundle is reserved for a later dated repository release and contains:

\begin{itemize}[nosep]
    \item \textbf{Unified corpus}: a 69-paper machine-readable inventory and a 92-record deduplicated screening ledger with inclusion/exclusion reasons and query protocol.
    \item \textbf{Diagnostic audit}: a frozen coding rubric, two independent blind ledgers, the unreconciled disagreement report, and source-located consensus fields for the 42-paper subset.
    \item \textbf{Atomic reporting audit}: source-located comparator, budget-parity, ablation, uncertainty, overhead, auxiliary-supervision, and replay-provenance fields for the same 42 papers.
    \item \textbf{CA-ID Card}: a blank machine-readable template for declaring the unit, estimand, effective state, intervention provenance, downstream protocol, support, uncertainty, and optimization interface.
    \item \textbf{Version manifest}: a dated snapshot note that distinguishes the frozen source bundle from later living-catalog updates.
\end{itemize}

\noindent The paper is the dated analytical snapshot; once published, each frozen bundle release should remain immutable while the repository continues as the evolving discovery and decision aid. Later releases should preserve prior versions and distinguish post-cutoff additions from corrections to frozen rows.

\section{Conclusion}
\label{sec:conclusion}

This paper has provided a dedicated survey of credit assignment in reinforcement learning for large language models, tracing the evolution from reasoning RL to agentic RL and identifying the fundamental challenges that drive methodological innovation.

Our analysis yields five key takeaways (annotated with evidence levels: \textbf{[SE]} = strong empirical, \textbf{[LS]} = limited but suggestive, \textbf{[AS]} = authors' synthesis):

\begin{enumerate}[nosep]
    \item \textbf{Credit assignment is a central challenge of LLM RL} \emph{[SE]}, and its importance grows as we move from reasoning to agentic settings. The shift from single-generation trajectories ($\sim$1K--30K tokens) to multi-turn agent interactions ($\sim$100K--1M tokens) transforms credit assignment from an optimization convenience into a training necessity.

    \item \textbf{Reasoning RL offers several tested CA families} \emph{[LS]}. Token-, segment-, and step-level methods report improvements within their respective studies when prefixes or effective states are branchable and verifiers are available. Their transfer depends on replay fidelity, checkpointability, and matched budgets.

    \item \textbf{Agentic RL exposes additional identification breaks} \emph{[LS]}. Hidden environment state, limited restoration, heterogeneous units, long horizons, and weak local verification motivate retrospective, hierarchical, and typed-action methods. Their proxy signals require calibration rather than causal interpretation by default.

    \item \textbf{LLM-as-Critic appears to be a distinctive paradigm} \emph{[LS]} not directly mirrored in classical RL. The ability to use LLMs for semantic evaluation of intermediate states (CAPO, SWEET-RL, LaRe, HCAPO, CriticSearch) opens a methodological axis that appears specific to the LLM era. Whether this approach will prove more effective than traditional value-based methods remains an open empirical question.

    \item \textbf{The field is accelerating} \emph{[AS---bibliometric observation]}. The July 31, 2026 corpus contains 69 papers (56 core and 13 adjacent/boundary) selected from 92 deduplicated records. This growth makes frozen cutoffs, machine-readable updates, and denominator-specific claims necessary rather than optional.
\end{enumerate}

As LLMs evolve from reasoning engines to autonomous agents operating in real environments, the question of credit assignment transforms from ``which reasoning step was correct?'' to ``which intervention changed the world under which downstream protocol?'' Beyond the original long-form taxonomy, the restored-state boundary, hidden-state witness, CA-ID Card, structured corpus, and source-located audit (\Cref{sec:identification,app:inventory,app:checklist,sec:release}) provide reusable tools for asking that sharper question without overstating cross-paper comparability.

\begin{takebox}[The Central Thesis of This Survey \emph{--- Evidence Level: Authors' Synthesis}]
\emph{We hypothesize that the shift from reasoning to agentic RL is not merely an extension of existing methods to harder tasks---it reshapes which credit claims are identifiable.} Reasoning CA more often benefits from prefix/effective-state closure, accessible branch points, local verification, and shorter horizons; agentic CA more often breaks those conditions through hidden environment state, irreversible actions, weak local verification, and long interaction horizons. The six diagnostics and CA-ID Card make these claims conditional and testable rather than treating ``reasoning'' and ``agentic'' as deterministic categories.
\end{takebox}

\FloatBarrier
\clearpage
\appendix

\section{Method Quick-Reference Index}
\label{app:index}

\Cref{tab:index} preserves 44 named methods from the original long-form narrative index, with full names and section references. Three legacy inventory rows not listed in that table (Turn-Level Reward, PRS, and AdaptSeg), together with the frontier additions that expand the unified corpus to 69 papers, appear in \Cref{tab:inventory}; their machine-readable frozen snapshot is reserved for the dated companion-repository release.

\begin{table}[h]
\centering
\caption{Alphabetical index of credit assignment methods reviewed in this paper.}
\label{tab:index}
\resizebox{\textwidth}{!}{
\begin{tabular}{llll}
\toprule
\textbf{Abbreviation} & \textbf{Full Name} & \textbf{Reference} & \textbf{Section} \\
\midrule
ACPO & Attribution-based Credit for RLVR & \citet{acpo} & \S\ref{sec:reasoning_step} \\
AgentPRM & Process Reward Model for LLM Agents & \citet{agentprm} & \S\ref{sec:agentic_prm} \\
ArCHer & Actor-Critic with Hierarchical Evaluation & \citet{archer} & \S\ref{sec:agentic_hierarchical} \\
C3 & Contextual Counterfactual Credit & \citet{c3} & \S\ref{sec:agentic_hindsight} \\
CAPO & Credit Assignment Policy Optimization & \citet{capo} & \S\ref{sec:reasoning_step} \\
CARL & Critical Action Reinforcement Learning & \citet{carl} & \S\ref{sec:agentic_hierarchical} \\
CCPO & Counterfactual Credit Policy Optimization & \citet{ccpo} & \S\ref{sec:agentic_hindsight} \\
CriticSearch & Retrospective Critic for Search Agents & \citet{criticsearch} & \S\ref{sec:agentic_hindsight} \\
Dr.\ MAS & Stable RL for Multi-Agent LLMs & \citet{drmas} & \S\ref{sec:multi_agent} \\
FinePO & Fine-Grained Process Reward (SketchVL) & \citet{fineprm} & \S\ref{sec:reasoning_step} \\
From $r$ to $Q^*$ & Implicit Token-Level Credit via DPO & \citet{fromrToQ} & \S\ref{sec:token_methods} \\
GiGPO & Group-in-Group Policy Optimization & \citet{gigpo} & \S\ref{sec:agentic_criticfree} \\
HCAPO & Hindsight Credit Assignment PO & \citet{hcapo} & \S\ref{sec:agentic_hindsight} \\
HICRA & Hierarchy-Aware Credit Assignment & \citet{hicra} & \S\ref{sec:reasoning_step} \\
IGPO & Information Gain Policy Optimization & \citet{igpo} & \S\ref{sec:agentic_info} \\
InT & Self-Proposed Interventions for CA & \citet{int} & \S\ref{sec:reasoning_step} \\
iStar & Implicit Step Rewards & \citet{istar} & \S\ref{sec:agentic_implicit} \\
ITPO & Implicit Turn-Level Process Rewards & \citet{itpo} & \S\ref{sec:agentic_prm} \\
LaRe & Latent Reward & \citet{lare} & \S\ref{sec:agentic_infra} \\
Lightning & Agent Lightning / LightningRL & \citet{agentlightning} & \S\ref{sec:agentic_infra} \\
LLM-MCA & LLM-based Multi-Agent CA & \citet{llmmca} & \S\ref{sec:multi_agent} \\
M-GRPO & Multi-Agent GRPO & \citet{mgrpo} & \S\ref{sec:multi_agent} \\
MAPPA & Multiagent Per-Action Process Awards & \citet{mappa} & \S\ref{sec:multi_agent} \\
PilotRL & Global Planning-Guided Progressive RL & \citet{pilotrl} & \S\ref{sec:agentic_hierarchical} \\
POAD & Policy Optimization with Action Decomposition & \citet{poad} & \S\ref{sec:agentic_criticfree} \\
PRL & Process Reward Learning & \citet{prl} & \S\ref{sec:reasoning_step} \\
PURE & Min-Form Process Reward & \citet{pure} & \S\ref{sec:reasoning_step} \\
QLLM & LLM-Generated Credit Functions & \citet{qllm} & \S\ref{sec:multi_agent} \\
RAGEN/StarPO & Star Policy Optimization & \citet{ragen} & \S\ref{sec:agentic_infra} \\
RED & Reward Redistribution to Token Level & \citet{red} & \S\ref{sec:token_methods} \\
SCAR & Shapley Credit Assignment Rewards & \citet{scar} & \S\ref{sec:segment_methods} \\
SHARP & Shapley Credit-based Multi-Agent Optimization & \citet{sharp} & \S\ref{sec:multi_agent} \\
SCRIBE & Structured Mid-Level Supervision & \citet{scribe} & \S\ref{sec:agentic_infra} \\
SPA-RL & Stepwise Progress Attribution & \citet{sparl} & \S\ref{sec:agentic_infra} \\
SPO & Segment Policy Optimization & \citet{spo} & \S\ref{sec:segment_methods} \\
SPRO & Self-Guided Process Reward & \citet{spro} & \S\ref{sec:reasoning_step} \\
SORL & Stabilizing Off-Policy RL (SO-PPO/SO-GRPO) & \citet{stppo} & \S\ref{sec:agentic_prm} \\
StepAgent & Step-Wise IRL Agent & \citet{stepagent} & \S\ref{sec:agentic_implicit} \\
SWEET-RL & Privileged Critic for Multi-Turn Agents & \citet{sweetrl} & \S\ref{sec:agentic_prm} \\
TARL & Turn-Level Adjudicated RL & \citet{tarl} & \S\ref{sec:agentic_prm} \\
TEMPO & Tree-Structured Credit Assignment & \citet{tempo} & \S\ref{sec:segment_methods} \\
T-REG & Token-Level Reward Regularization & \citet{treg} & \S\ref{sec:token_methods} \\
Turn-PPO & Turn-Level Optimized PPO & \citet{turnppo} & \S\ref{sec:agentic_prm} \\
VinePPO & Monte Carlo Token-Level PPO & \citet{vineppo} & \S\ref{sec:token_methods} \\
\bottomrule
\end{tabular}
}
\end{table}

\section{Complete Paper Inventory}
\label{app:inventory}

\Cref{tab:inventory} provides the 69-paper snapshot with taxonomy labels and compact metadata. \textbf{Type}: C = Core CA method, E = adjacent/boundary enabler. \textbf{Setting}: R = Reasoning RL, A = Agentic RL, M = Multi-Agent. \textbf{BL}: Baseline family: G = GRPO, P = PPO, D = DPO, O = ORM, T = TD. \textbf{Ev.}: \textbf{S} = strong empirical, \textbf{L} = limited but suggestive, \textbf{A} = primarily analytical, and \textbf{NC} = not coded in the fixed 42-paper full-text subset. The source-located CSV is authoritative when this compact table omits detail.

\begin{table}[h]
\centering
\caption{Unified inventory, Part I: original long-form rows 1--47 (39 core and 8 adjacent/boundary after consistent reclassification).}
\label{tab:inventory}
\renewcommand{\arraystretch}{0.93}
\resizebox{\textwidth}{!}{
\begin{tabular}{rlccccccl}
\toprule
\textbf{\#} & \textbf{Method} & \textbf{Type} & \textbf{Setting} & \textbf{Gran.} & \textbf{Methodology} & \textbf{BL} & \textbf{Ev.} & \textbf{Primary Benchmarks} \\
\midrule
\multicolumn{9}{l}{\textit{Reasoning RL --- Core CA Methods (15)}} \\
\midrule
1 & VinePPO & C & R & Token & MC & P & S & GSM8K, MATH \\
2 & RED & C & R & Token & Redistribution & P & L & MATH \\
3 & T-REG & C & R & Token & Self-generated & P & L & GSM8K, MATH \\
4 & From $r$ to $Q^*$ & C & R & Token & Implicit & D & A & Theoretical analysis \\
5 & SPO & C & R & Segment & MC & G & S & MATH-500, GSM8K \\
6 & SCAR & C & R & Segment & Game-theoretic & G & L & MATH \\
7 & TEMPO & C & R & Token/Seg & Tree-TD & P & L & MATH, GSM8K \\
8 & PURE & C & R & Step & Min-form PRM & G & S & MATH-500, AIME'24 \\
9 & SPRO & C & R & Step & Masked Adv. & G & S & MATH-500, AMC \\
10 & CAPO & C & R & Step & LLM-as-Critic & G & S & MATH-500, AIME'24 \\
11 & ACPO & C & R & Step & Attribution & G & L & MATH \\
12 & HICRA & C & R & Step & Hierarchy & G & S & AIME'24, AIME'25 \\
13 & PRL & C & R & Step & Entropy-RL & G & L & MATH, GSM8K \\
14 & InT & C & R & Step & Intervention & G & L & MATH \\
15 & FinePO & C & R & Sub-step & Fine PRM & --- & L & Domain-specific (visual) \\
\midrule
\multicolumn{9}{l}{\textit{Agentic RL --- Core CA Methods (20)}} \\
\midrule
16 & ArCHer & C & A & Turn & TD (hierarchical) & T & S & Multi-turn dialogue \\
17 & StepAgent & C & A & Step & Implicit+IRL & G & L & Tool-use tasks \\
18 & POAD & C & A & Token/Turn & Action Decomp. & P & S & Interactive tasks \\
19 & GiGPO & C & A & Step & MC (group) & G & S & ALFWorld, WebShop \\
20 & SWEET-RL & C & A & Turn & Privileged Critic & D & S & ColBench Backend \\
21 & AgentPRM & C & A & Step & TD+GAE & O & S & WebShop, TextCraft \\
22 & Turn-Level & C & A & Turn & Hybrid & G & L & Web navigation \\
23 & Turn-PPO & C & A & Turn & Turn-level MDP & G & S & WebShop \\
24 & SORL & C & A & Turn & Bias-corrected & G & L & Multi-turn search \\
25 & TARL & C & A & Turn & LLM-Judge & G & S & $\tau$-bench \\
26 & ITPO & C & A & Turn & Implicit & D & L & Dialogue tasks \\
27 & IGPO & C & A & Turn & Info-theoretic & G & L & Agentic tasks \\
28 & CARL & C & A & Step & Entropy-based & G & S & HotpotQA, 2WikiMQA \\
29 & iStar & C & A & Step & Implicit DPO & D & L & Trajectory pairs \\
30 & PilotRL & C & A & Step & Progressive & G & L & Agentic planning \\
31 & LaRe & C & A & Step & LLM-Critic & G & L & Symbolic + agentic \\
32 & HCAPO & C & A & Turn & Hindsight & G & S & Agentic tasks \\
33 & C3 & C & A/M & Turn & Counterfactual & G & L & Multi-agent + agentic \\
34 & CCPO & C & A/M & Turn & Counterfactual & G & L & Agentic tasks \\
35 & CriticSearch & C & A & Turn & Retrospective Critic & G & S & Multi-hop QA \\
\midrule
\multicolumn{9}{l}{\textit{Agentic RL --- CA-Adjacent Enablers (6)}} \\
\midrule
36 & SPA-RL & E & A & Step & MLP estimator & G & L & Agentic tasks \\
37 & Lightning & E & A & Step & Decoupled Arch. & G & L & Multi-turn agents \\
38 & RAGEN & E & A & Step & Uncertainty & G & S & Benchmark suite \\
39 & SCRIBE & E & A & Step & Skill-prototype & G & L & Agentic tasks \\
40 & PRS & E & A & Step & Progressive & G & S & Progressive tasks \\
41 & AdaptSeg & E & A & Segment & Segmentation & G & L & Agentic tasks \\
\midrule
\multicolumn{9}{l}{\textit{Multi-Agent --- Core and Adjacent/Boundary Methods (6)}} \\
\midrule
42 & M-GRPO & C & M & Multi-Agent & Hierarchical & G & L & Multi-agent tasks \\
43 & LLM-MCA & E & M & Multi-Agent & LLM-Critic & G & L & Multi-agent eval \\
44 & QLLM & E & M & Multi-Agent & LLM-generated & G & L & Multi-agent tasks \\
45 & SHARP & C & M & Multi-Agent & Shapley & G & S & Multi-agent tasks \\
46 & MAPPA & C & M & Multi-Agent & Per-action PRM & G & S & AIME, AMC \\
47 & Dr.\ MAS & C & M & Multi-Agent & Agent-wise Adv. & G & S & Math tasks \\
\bottomrule
\end{tabular}
}
\end{table}

\begin{table}[p]
\centering
\caption{Unified inventory, Part II: rows 48--50 complete the fixed 42-core-paper audit; rows 51--69 are frontier additions outside that full-text audit. Parts I--II total 56 core and 13 adjacent/boundary papers.}
\label{tab:inventory_frontier}
\resizebox{\textwidth}{!}{
\begin{tabular}{rlccccccl}
\toprule
\textbf{\#} & \textbf{Method} & \textbf{Type} & \textbf{Setting} & \textbf{Gran.} & \textbf{Methodology} & \textbf{BL} & \textbf{Ev.} & \textbf{Primary Benchmarks} \\
\midrule
\multicolumn{9}{l}{\textit{May-to-July additions integrated into the unified snapshot (22)}} \\
\midrule
48 & GVPO & C & A & Step/Turn & Verifier shaping & G & S & AppWorld agents \\
49 & A$^3$ & C & A & Turn/Action & Structured Adv. & G & S & ShellOps CLI \\
50 & A$^2$TGPO & C & A & Turn & IG/Clipping & G & L & Tool agents \\
51 & AutoResearch & E & A & Trial & Lineage feedback & --- & NC & Research agents \\
52 & Memory-R2 & C & A & Memory op. & Matched-state groups & --- & NC & Memory agents \\
53 & GRPO-$\lambda$ & C & R & Token/Seq. & $\lambda$-return & G & NC & Math reasoning \\
54 & DelTA & C & R & Token & Discriminative weight & G & NC & RLVR reasoning \\
55 & SCRL & C & R & Span & Subproblem curriculum & G & NC & Math reasoning \\
56 & TRIAGE & C & A & Segment/Turn & Role-typed credit & G & NC & Agentic tasks \\
57 & OAR & C & R & Token/Step & Outcome sensitivity & G & NC & Math reasoning \\
58 & CRAFT & C & A & Token/Step & Sibling rollouts & G & NC & Agentic tasks \\
59 & SC-GRPO & C & R & Token & Self-conditioned KL & G & NC & RLVR reasoning \\
60 & GRAIL & C & R & Token & Gradient saliency & G & NC & RLVR reasoning \\
61 & VPR & C & A & Turn & Verifiable process & G & NC & Agentic reasoning \\
62 & APPO & C & A & Decision/Proc. & Procedural branch & G & NC & Agentic tasks \\
63 & OPID & E & A & Token/Skill & Skill distillation & G & NC & Agentic tasks \\
64 & PAPO & E & R & Token/Entropy & Entropy polarity & G & NC & RLVR reasoning \\
65 & AT$^2$PO & C & A & Turn & Tree advantage & G & NC & Multi-turn agents \\
66 & AEM & C & A & Response/Turn & Entropy modulation & G & NC & Multi-turn agents \\
67 & T$^2$PO & E & A & Token/Turn & Exploration control & G & NC & Multi-turn agents \\
68 & Rubric-Grounded RL & E & R & Criterion/Step & Judge shaping & G & NC & Reasoning \\
69 & PivoARL & C & A & Turn/Action & Pivotal retry & G & NC & Agentic tasks \\
\midrule
\multicolumn{9}{l}{\textit{Background / Foundational (not counted in 69)}} \\
\midrule
& Math-Shepherd & --- & R & Step & MC labeling & --- & S & GSM8K, MATH \\
& OmegaPRM & --- & R & Step & MC labeling & --- & S & MATH \\
& GRPO & --- & R & Episode & Group baseline & --- & S & Math, code \\
& DeepSeek-R1 & --- & R & Episode & GRPO & --- & S & AIME, MATH, code \\
\bottomrule
\end{tabular}
}
\end{table}

\paragraph{Note on classification and denominators.} Part I contains 39 core and 8 adjacent/boundary rows after LLM-MCA and QLLM are consistently coded as boundary enablers. GVPO, A$^3$, and A$^2$TGPO (rows 48--50) are the remaining three core methods in the fixed 42-paper audit, with evidence labels S, S, and L. Only rows 51--69 carry \textbf{NC}: they belong to the 69-paper qualitative corpus but not the 42-paper full-text diagnostic/reporting subset. Integrating all rows yields 56 core and 13 adjacent/boundary papers. \textbf{NC} is a denominator marker, not a weak-evidence rating.

\section{Reporting Checklist for Future Credit Assignment Papers}
\label{app:checklist}

Based on the methodological gaps identified in this survey, we propose the following reporting checklist for future CA papers.

\begin{table}[h]
\centering
\caption{Recommended reporting checklist for credit assignment papers in LLM RL.}
\label{tab:checklist}
\resizebox{\textwidth}{!}{
\begin{tabular}{llp{7.5cm}}
\toprule
\textbf{Category} & \textbf{Priority} & \textbf{Item} \\
\midrule
\multirow{2}{*}{Model \& Data} & Required & Base model name, size, and version \\
& Required & Training data: source, size, and any filtering applied \\
\midrule
\multirow{2}{*}{CA Method} & Required & Credit granularity (token / segment / step / turn / multi-agent) \\
& Required & Methodology family per our taxonomy \\
\midrule
\multirow{3}{*}{Baselines} & Required & At least one episode-level baseline (GRPO or PPO) with identical base model \\
& Required & Baseline training recipe: same compute budget or explicit compute comparison \\
& Recommended & CA-component ablation isolating the contribution \\
\midrule
\multirow{3}{*}{Evaluation} & Required & Benchmark names and specific splits \\
& Required & Evaluation metric with exact definition \\
& Recommended & Variance estimates (std across $\geq$3 seeds, or confidence intervals) \\
\midrule
\multirow{2}{*}{Compute} & Required & Total GPU-hours for training \\
& Recommended & CA-specific overhead: additional forward passes, environment resets, or LLM calls \\
\midrule
\multirow{2}{*}{Trajectory Info} & Required & Average trajectory length (tokens for reasoning, turns for agentic) \\
& Recommended & Trajectory length distribution (min, median, max) \\
\bottomrule
\end{tabular}
}
\end{table}

\paragraph{Reliability of the diagnostic coding.} Rather than validating the checklist on three hand-picked papers, we independently and blindly recoded all 42 papers in the full-text subset under rules frozen before either pass. \Cref{tab:checklist_validation} reports exact agreement and chance-corrected agreement; all disagreements are retained in the frozen coding ledgers planned for the dated companion-repository release.

\begin{table}[h]
\centering
\caption{Independent blind cross-coding reliability in the fixed 42-paper subset.}
\label{tab:checklist_validation}
\resizebox{\textwidth}{!}{
\begin{tabular}{lrrrr}
\toprule
\textbf{Diagnostic} & \textbf{Agreement} & \textbf{Percent} & \textbf{Cohen's $\kappa$} & \textbf{Disagreements} \\
\midrule
Transition non-closure & 39/42 & 92.9\% & .856 & 3 \\
Partial observability & 33/42 & 78.6\% & .557 & 9 \\
Limited replay & 37/42 & 88.1\% & .762 & 5 \\
Action heterogeneity & 37/42 & 88.1\% & .759 & 5 \\
Weak local verifiability & 36/42 & 85.7\% & .543 & 6 \\
Agent coupling & 41/42 & 97.6\% & .909 & 1 \\
Exclusive principal family & 42/42 & 100.0\% & 1.000 & 0 \\
\bottomrule
\end{tabular}
}
\end{table}

\paragraph{Audited reporting gaps.} Across 252 diagnostic cells, binary agreement is 223/252 (88.5\%). The separate source-located reporting audit finds 40/42 matched episode/trajectory comparators and 39/42 CA-specific ablations, but only 19/42 full model/data/budget parity. CA-specific overhead is numeric for 21, a numeric proxy for 2, qualitative for 12, not reported for 6, and not applicable for 1. Uncertainty is based on repeated training in 10 papers; 7 repeat evaluation only, 1 uses evaluation bootstrap only, 1 reports a significance test only, 2 are unclear, and 21 do not report uncertainty. These denominated fields replace the earlier informal claims about all papers.


\FloatBarrier
\clearpage
\bibliographystyle{plainnat}
\bibliography{references}

\end{document}